\documentclass[11pt,a4paper]{article}
\usepackage[hyperref]{acl2021}
\usepackage{times}
\usepackage{latexsym}
\usepackage[linesnumbered,ruled]{algorithm2e}
\usepackage{subfig}
\usepackage{graphicx}
\usepackage[T1]{fontenc}
\usepackage{graphics}
\usepackage{multirow}
\usepackage{multicol}
\usepackage{enumitem}
\usepackage{soul,xcolor}
\usepackage{amsmath}
\usepackage{enumitem}
\usepackage{colortbl}
\usepackage{arydshln}
\usepackage{microtype}

\aclfinalcopy 
\def\aclpaperid{3684} 

\usepackage{tikz}
 
\usepackage{pifont}

\newcommand{\name}{\textsf{HIT}}

\title{\name: A Hierarchically Fused Deep Attention Network for Robust Code-mixed Language Representation}

\author{Ayan Sengupta$^1$, Sourabh Kumar Bhattacharjee$^1$, \\ \textbf{Tanmoy Chakraborty$^2$, Md Shad Akhtar$^2$} \\
  $^1$Optum Global Advantage (UnitedHealth Group), Noida, India\\ 
$^2$Dept. of CSE, IIIT-Delhi, India\\
  \texttt{\{ayan.sengupta007, sourabhb398\}@gmail.com}; \\
  \texttt{\{tanmoy, shad.akhtar\}@iiitd.ac.in} \\
  }

\date{}

\begin{document}
\maketitle
\begin{abstract}

Understanding linguistics and morphology of resource-scarce code-mixed texts remains a key challenge in text processing. Although word embedding comes in handy to support downstream tasks for low-resource languages, there are plenty of scopes in improving the quality of language representation particularly for code-mixed languages. In this paper, we propose \name, a robust representation learning method for code-mixed texts. \name\ is a hierarchical transformer-based framework that captures the semantic relationship among words and hierarchically learns the sentence-level semantics using a fused attention mechanism. \name\ incorporates two attention modules, a multi-headed self-attention and an outer product attention module, and computes their weighted sum to obtain the attention weights. Our evaluation of \name\ on one European (\textit{Spanish}) and five Indic (\textit{Hindi}, \textit{Bengali}, \textit{Tamil}, \textit{Telugu}, and \textit{Malayalam}) languages across four NLP tasks on eleven datasets suggests significant performance improvement against various state-of-the-art systems. We further show the adaptability of learned representation across tasks in a transfer learning setup (with and without fine-tuning).



\end{abstract}

\section{Introduction}

India is the second most populated country in the world, where $\sim1.36$ billion people speak in over $200$ different languages. Among them, the top five languages ({\em Hindi, Bengali, Telegu, Tamil,} and {\em Malayalam}) covers $\sim 93\%$
of the entire population with more than $26\%$ of them being bilingual (as per Wikipedia).
Moreover, a significant proportion of them ~\cite{singh-etal-2018-language} use code-mixed languages to express themselves in Online Social Networks (OSN).

\emph{Code-mixing} (CM) is a linguistic phenomenon in which two or more languages are alternately used during conversation. One of the languages is usually English, while the other can be any regional language such as Hindi (Hindi + English $\rightarrow$ \textit{Hinglish}), Bengali (Bengali + English $\rightarrow$ \textit{Benglish}), Spanish (Spanish + English $\rightarrow$ \textit{Spaniglish}), etc.
Their presence on social media platforms and in day-to-day conversions among the people of a multi-lingual communities (such as Indians) is overwhelming. Despite the fact that a significant population is comfortable with code-mixed languages, the research involving them is fairly young. One of the prime reasons is the linguistic diversity, i.e., research on any language often fails to adapt for other distant languages, thus they need to be studied and researched separately. In recent years, many  organizations have identified the challenges and have put in commendable efforts for the development of computational systems in regional monolingual or code-mixed setups.

Traditionally, the NLP community has studied the code-mixing phenomenon from a task-specific point of view. Recently, a few studies \cite{pratapa-etal-2018-word, aguilar-solorio-2020-english} have started learning representations for code-mixed texts for semantic and syntactic tasks. While the former has showcased the importance of multi-lingual embeddings from CM text, the latter has made use of a hierarchical attention mechanism on top of positionally aware character bi-grams and tri-grams to learn robust word representations for CM text. Carrying over the same objective, in this paper, we introduce a novel \textbf{HI}erarchically attentive \textbf{T}ransformer (\name) framework to effectively encode the syntactic and semantic  features in embeddings space. At first, \name\ learns sub-word level representations employing a fused attention mechanism (FAME) -- an \textit{outer-product} based attention mechanism \cite{pmlr-v119-le20b} fused with standard multi-headed self-attention \cite{vaswanietal2017}. The intuition of sub-word level representation learning is supplemented by the lexical variations of a word in code-mixed languages. The \textit{character-level} \name\ helps in representing phonetically similar word and their variations to a similar embedding space and extracts better representation for noisy texts. Subsequently, we apply \name\ module at \textit{word-level} to incorporate the semantics at the sentence-level. The output of \name\ is a sequence of word representations, and can be fed to the architectures of any downstream NLP tasks. For the evaluation of \name, we experiment on one classification (sentiment classification), one generative (MT), and two sequence-labelling (POS tagging and NER) tasks. In total, these tasks span to eleven datasets across six code-mixed languages -- one European (\textit{Spanish}) and five Indic (\textit{Hindi}, \textit{Bengali}, \textit{Telugu}, \textit{Tamil}, and \textit{Malayalam}).   
Our empirical results show that representations learned by \name\ are superior to existing multi-lingual and code-mixed representations, and report state-of-the-art performance across all tasks. Additionally, we  observe encouraging adaptability of \name\ in a transfer learning setup across tasks. The representations learned for a task is employed for learning other tasks w/ and w/o fine-tuning. \name\ yields good performance in both setups for two code-mixed languages. 


\noindent \textbf{Main contributions:} We summarize our contributions as follow:
\begin{itemize}[leftmargin=*,topsep=0pt,itemsep=-1ex,partopsep=1ex,parsep=1ex]
    \item We propose a hierarchical attention transformer framework for learning word representations of code-mixed texts for six non-English languages. 
    
    \item We propose a hybrid self-attention mechanism, FAME, to fuse the multi-headed self-attention and outer-product attention mechanisms in our transformer encoders.
    
    \item We show the effectiveness of \name\ on eleven datasets across four NLP tasks and six languages.
    
    \item We observe good task-invariant performance of \name\ in a transfer learning setup for two code-mixed languages.
    
\end{itemize}

\noindent \textbf{Reproducibility:} Source codes, datasets and other details to reproduce the results have been made public at \url{https://github.com/LCS2-IIITD/HIT-ACL2021-Codemixed-Representation}. 
\begin{figure*}[hbt!]
\centering
\includegraphics[scale=.65]{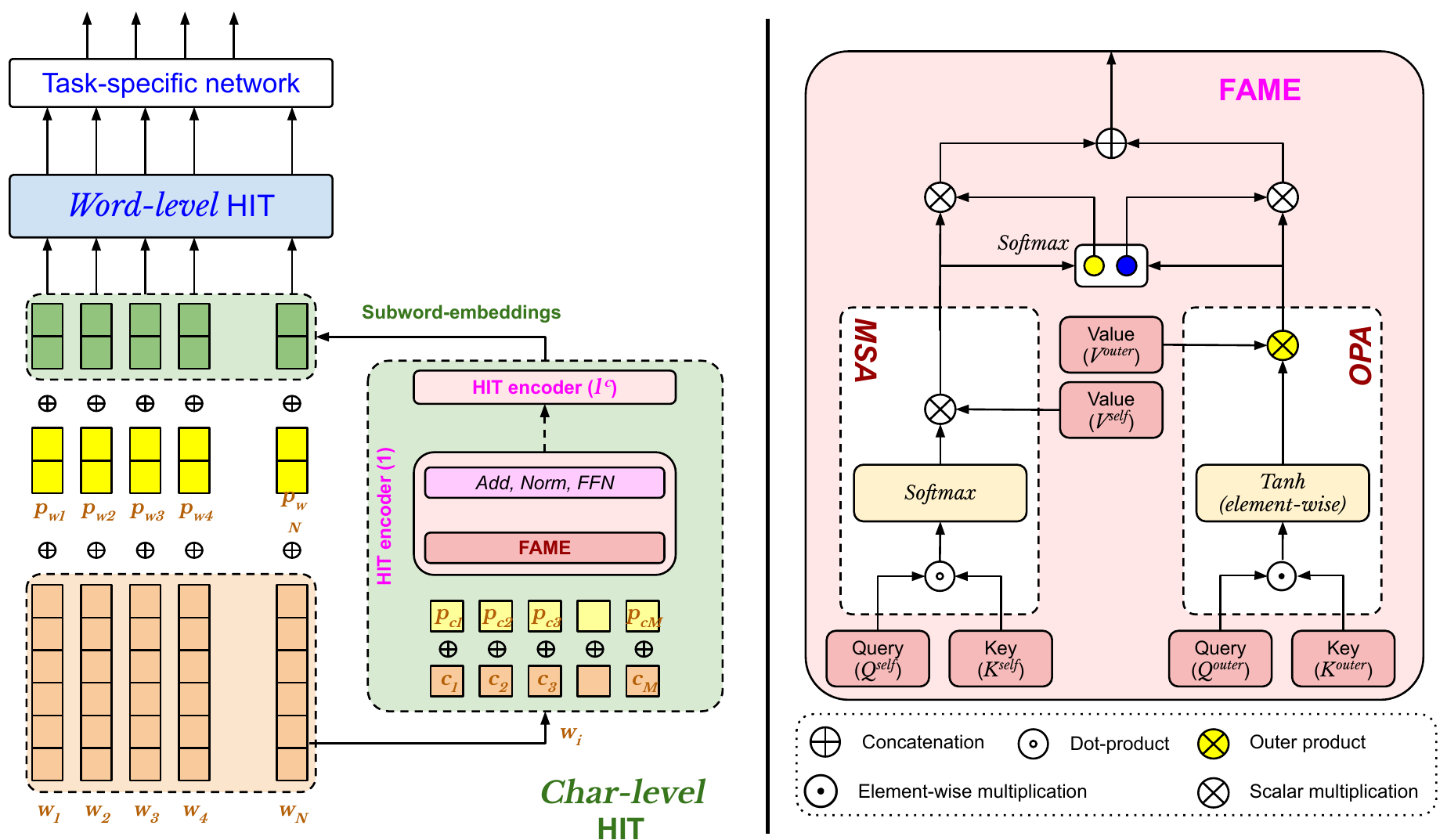}
\caption{\textbf{Hi}erarchical \textbf{T}ransformer along with our novel {\bf FAME} mechanism for attention computation.}
\label{fig:model}
\end{figure*}
\section{Related Work}
Recent years have witnessed a few interesting work in the domain of code-mixed/switched representation learning. Seminal work was driven by bilingual embedding that employs cross-lingual transfer to develop NLP models for resource-scarce languages \cite{upadhyay-etal-2016-cross, akhtar-etal-2018-solving,10.1613/jair.1.11640}. \newcite{faruqui-dyer-2014-improving} introduced the BiCCA embedding using bilingual correlation, which performed well on syntactical tasks, but poorly on cross-lingual semantic tasks. Similarly, frameworks proposed by \newcite{hermann-blunsom-2014-multilingual} and \newcite{luong-etal-2015-bilingual} depend on projecting the words of two languages into a single embedding space.

However, as demonstrated by \citet{pratapa-etal-2018-word}, bilingual embedding techniques are not ideal for CS text processing and should be replaced by multi-lingual embeddings learnt from CM data. The transformer-based Multilingual BERT \cite{devlin-etal-2019-bert} embedding has been demonstrated \cite{pires-etal-2019-multilingual} to possess impressive cross-lingual model transfer capabilities. 
Also, the XLM model \cite{NEURIPS2019_c04c19c2} has also shown the effects of cross-lingual training for low-resource and CM language tasks.

Another school of thought revolves around sub-word level representations, which can help to capture variations found in CM and transliterated text. \newcite{joshi2016towards} proposed a CNN-LSTM based model to learn the sub-word embeddings through 1-D convolutions of character inputs. They showed that it resulted in better sentiment classification performance for CM text. On top of this intuition, attention-based frameworks have also been proven to be successful in learning low-level representations. The HAN \cite{yang2016hierarchical} model provides the intuition of hierarchical attention for document classification, which enables it to differentially attend to more and less important content, at the word and sentence levels. 
In another work, \newcite{aguilar-solorio-2020-english} proposed CS-ELMo for code-mixed inputs with similar intuition.
It utilizes the hierarchical attention mechanism on bi-gram and tri-gram levels to effectively encode the sub-word level representations, while adding positional awareness to it.

Our work builds on top of these two earlier works to push the robustness of code-mixed representations to higher levels. However, the main difference between existing studies and \name\ is the incorporation of outer-product attention-based fused attention mechanism (FAME).
\section{Proposed Methodology}
In this section, we describe the architecture of \name\ for learning effective representations in code-mixed languages. 
The backbone of \name\ is transformer \cite{vaswanietal2017} and  Hierarchical Attention Network (HAN) \cite{yang2016hierarchical}. \name\ takes a sequence of words (a code-mixed sentence) $S = \langle w_1, w_2, ..., w_N\rangle$ as input and processes each word $w_i$ using a \textit{character-level} \name\ to obtain sub-word representation $S^{sb} = \langle sb_1, sb_2, ..., sb_N\rangle $. The \textit{character-level} \name\ is a transformer encoder, where instead of computing multi-headed self-attention only, we amalgamate it with an outer-product attention mechanism \cite{pmlr-v119-le20b} as well. The intuition of outer-product attention is to extract higher-order character-level relational similarities among inputs. To leverage both attention mechanisms, we compute their weighted sum using a softmax layer. Subsequently, we pass it through the typical \textit{normalization} and \textit{feed-forward} layers to obtain the encoder's output. A stacking of $l_c$ encoders is used. In the next layer of the hierarchy, these sub-word representations are combined with positional and rudimentary embeddings of each word and forwarded to the \textit{word-level} \name's encoder. Finally, the output of \textit{word-level} \name\ is fed to the respective task-specific network.
The hierarchical nature of \name\ enables us to capture both \textit{character-level} and \textit{word-level} relational (syntactic and semantic) similarities. A high-level schema of \name\ is shown in Figure \ref{fig:model}.

\subsection{Fused Attention Mechanism (FAME)}
\label{subs: FAME}
FAME extends the multi-headed self-attention (MSA) module of a standard transformer by including a novel outer-product attention (OPA) mechanism. Given an input $x$, we use three weight matrices, $W^{self}_Q, W^{self}_K,$ and $W^{self}_V$, to project the input to \textit{Query} ($Q^{self}$) , \textit{Key} ($K^{self}$), and \textit{Value} ($V^{self}$) representations for MSA, respectively. Similarly for OPA we use $W^{outer}_Q, W^{outer}_K,$ and $W^{outer}_V$ for the projecting $x$ to $Q^{outer}, K^{outer}$ and $V^{outer}$. Next, the two attention mechanisms are learnt in parallel, and a weighted sum is computed as its output. Formally, $ H = \alpha_1 \cdot Z_{self} \oplus \alpha_2 \cdot Z_{outer}$, 
where $Z_{self}$ and $Z_{outer}$ respectively are the outputs of multi-headed self attention and outer-product attention modules, and $\alpha_1$ and $\alpha_2$ are the respective weights computed through a softmax function.


\paragraph{Multi-Headed Self Attention.}
 
The standard transformer self-attention module \cite{vaswanietal2017} computes a scaled dot-product between the \textit{query} and \textit{key} vectors prior to learn the attention weights for the \textit{value} vector. We compute the output as follows:
\begin{eqnarray}\tiny
Z_{self} & = & softmax\left( \frac{Q^{self} \cdot K^{self^{T}}}{\sqrt{d^k}}\right) V^{self} \nonumber \\ \nonumber 
& = & \sum_i^N softmax\left( \frac{q \cdot k_i}{\sqrt{d^k}}\right) v_i , \forall q \in Q^{self}
\end{eqnarray}
where $N$ is the sequence length, and $d^k$ is the dimension of the \textit{key} vector.




\paragraph{Outer-Product Attention.}
\label{subsubs: OPA}
%
%
This is the second attention that we incorporate into FAME. Although the fundamental process of OPA \cite{pmlr-v119-le20b} is similar to the multi-headed self-attention computation, OPA differs in terms of different operators and the use of row-wise \textit{tanh} activation instead of softmax. To compute the interaction between the query and key vectors, we use element-wise multiplication as opposed to the dot-product in MSA. Subsequently, an element-wise \textit{tanh} is applied before computing the outer-product with the value vector. The intuition is to exploit fine-level associations between the \textit{key}-scaled \textit{query} and \textit{value} representations in a code-mixed setup. Similar to the earlier case, we define OPA as:
\begin{eqnarray}
Z_{outer} = \sum_i^N tanh\left(\frac{q \odot k_i}{\sqrt{d_k}}\right) \otimes v_i,  \forall q \in Q^{outer} \nonumber
\end{eqnarray}
where $\odot$ is the element-wise multiplication, and $\otimes$ is the outer-product.

\subsection{Task-specific Layers}
As we mention earlier, \name\ can be adapted for various NLP tasks including sequence labelling, classification, or generative problems. In the current work, we evaluate \name\ on part-of-speech (POS) tagging, named-entity recognition (NER), sentiment classification, and machine translation (MT). We mention their specific architectural details below.

    For the sentiment classification, we apply a {\em GlobalAveragePooling} operation over the token embeddings to obtain the sentence embeddings. Additionally, we concatenate extracted statistical features along with the embeddings before feeding into the final classification layer. We use \textit{tf-idf} (term frequency–inverse document frequency) vectors for $\{1,2,3\}$-grams of words and characters extracted from each text. We hypothesize that these statistical features contain sufficient information to get rid of any handcrafted features like the ones suggested by 
    ~\citet{bansal-etal-2020-code}. Finally, a \textit{softmax} activation function is used for the prediction.
    Similarly, for POS tagging and NER, the corresponding labels for each of the token's embedding is obtained through a softmax activated output.
    In case of MT, we use an encoder-decoder framework  where both the encoder and the decoder are based on the {\name} framework.
    


\section{Experiments, Results, and Analyses}
In this section, we furnish the details of chosen datasets, our experimental results, comparative study, and necessary analyses.

          

\begin{table}[t]
    \centering
    \resizebox{\columnwidth}{!}{
    \begin{tabular}{c|c|c|c|c|c|c|c}
         \multirow{2}{*}{\bf Tasks} & \multirow{2}{*}{\bf Lang} & \multicolumn{2}{c|}{\bf Train} & \multicolumn{2}{c|}{\bf Test} & \multirow{2}{*}{\bf Total} & \multirow{2}{*}{\bf \#Labels} \\ \cline{3-6}
          &  & \bf \#Sent & \bf \#Token & \bf \#Sent & \bf \#Token & & \\ \hline
          
          \hline
         \multirow{4}{*}{\bf POS} & Hi* & 1191 & 6575 & 148 & 2300 & 1489 & 14\\
          & Te* & 1,585 & 7,190 & 198 & 2,927 & 1,982 & 52\\
           & Be* & 500 & 4,108 & 62 & 631 & 626 & 39\\
            & Sp & 27,893 & 11,897 & 4,298 & 3,866 & 36,489 & 17\\ \hline 
         \multirow{2}{*}{\bf NER} & Hi* & 1663 & 9,397 & 207 & 3,272 & 2,079 & 7\\
          & Sp & 33,611 & 52,680 & 10,085 & 23,787 & 53,781 & 19\\ \hline
         \multirow{4}{*}{\bf Sentiment} & Hi* & 3,103 & 9,005 & 387 & 3,191 & 3,879 & 3\\
          & Ta & 11,335 & 27,476 & 3,149 & 10,339 & 15,744 & 4\\
           & Ma & 4,851 & 16,551 & 1,348 & 6,028 & 6,739 & 4\\
            & Sp & 12,194 & 28,274 & 1,859 & 7,822 & 15,912 & 3\\ \hline
        \multirow{2}{*}{\bf MT} & En (Src) & \multirow{2}{*}{248,330} & 84,609 & \multirow{2}{*}{2,000} & 5,314 & \multirow{2}{*}{252,330} & \multirow{2}{*}{-} \\
         & Hi (Tgt) & & 108,442 & & 5,797 & & \\
            \hline
    \end{tabular}}
    \caption{Dataset statistics. Star($^*$) signifies 90-10 ratio. 
    }
    \label{tab:dataset}
\end{table}
\subsection{Datasets}
\label{subs: data}
We evaluate 11 publicly available datasets across 4 tasks in 6 code-mixed languages. For POS tagging, we employ Hindi, Telugu, Bengali, and Spanish, whereas, we evaluate Hindi and Spanish datasets for NER. Similarly, in sentiment classification, we incorporate Hindi, Tamil, Malayalam, and Spanish code-mixed sentences. Finally, for machine translation, we use a recently released Hindi-English code-mixed parallel corpus. A brief statistics of all datasets is presented in Table \ref{tab:dataset}.\\ 
$\bullet$ \textbf{POS tagging:} We use the Hindi-English code-mixed POS dataset provided by ~\citet{singh-etal-2018-twitter}. It was collected from  Twitter and has $1489$ sentences. Each token in the sentence is tagged with one of the 14 tags\footnote{We furnish the details of tagset in the appendix}. The Bengali and Telugu datasets are collected from ICON-2016 workshop\footnote{\url{http://amitavadas.com/Code-Mixing.html}}. The instances are the social-media messages, collected from Twitter, Facebook and WhatsApp, and have $1982$ and $626$ sentences in Telugu and Bengali, respectively. These two datasets follow Google universal tagset \cite{DBLP:journals/corr/abs-1104-2086} and contain $52$ and $39$ tags respectively. 
For Spanish, we use Linguistic Code-switching Evaluation (LinCE) POS dataset ~\cite{alghamdi-etal-2016-part} consisting of more that $35k$ sentences with $14$ tags.

\begin{table*}
\centering
\resizebox{0.9\textwidth}{!}
{
\begin{tabular}{l|ccc|ccc|ccc||ccc}
\hline
 \multirow{2}{*}{\textbf{Model}} & \multicolumn{3}{c|}{\bf Hindi} & \multicolumn{3}{c|}{\bf Tamil} & \multicolumn{3}{c||}{\bf Malayalam} & \multicolumn{3}{c}{\bf Spanish} \\
  & \textbf{Pr.} & \textbf{Re.} & \textbf{F1} & \textbf{Pr.} & \textbf{Re.} & \textbf{F1} &  \textbf{Pr.} & \textbf{Re.} & \textbf{F1} &  \textbf{Pr.} & \textbf{Re.} & \textbf{F1}
  \\
\hline
BiLSTM & 0.916 & 0.901 & 0.909 & 0.502 & 0.428 & 0.451 & 0.653 & 0.588 & 0.612 & 0.429 & 0.431 & 0.428 \\
Subword-LSTM & 0.905 & 0.907 & 0.905 & 0.503 & 0.418 & 0.426 & 0.577 & 0.592 & 0.581 &  0.445 & 0.437 & 0.432 \\
HAN & 0.915 & 0.906 & 0.908 & 0.490 & 0.411 & 0.439 & 0.639 & 0.611 & 0.634 & 0.449 & 0.439 & 0.440 \\
ML-BERT & 0.919 & \textbf{0.914} & 0.909 & 0.260 & 0.310 & 0.280 & 0.600 & 0.630 & 0.610 & 0.451 & 0.419 & 0.437 \\
CS-ELMO  & 0.921 & 0.903 & 0.909 & 0.515 & 0.432 & 0.459 & 0.666 & 0.623 & 0.642 & 0.429 & 0.453 & 0.431\\
\hline
\rowcolor{blue!10} \textbf{\name} & \textbf{0.956} & 0.914 & \textbf{0.915} & 0.499 & \textbf{0.451} & \textbf{0.473} & 0.710 & 0.628 & 0.651 & \textbf{0.502} & \textbf{0.454} & \textbf{0.460} \\ \hline
(-) $Atn^{outer}$  & 0.933 & 0.911 & 0.913 & \textbf{0.520} & 0.448 & 0.455 & \textbf{0.718} & 0.624 & \textbf{0.655} & 0.463 & 0.440 & 0.445 \\ 
(-) \textit{char-level} \name & 0.903 & 0.887 & 0.901 & 0.504 & 0.418 & 0.432 & 0.659 & 0.605 & 0.627 & 0.448 & 0.438 & 0.433 \\ 
\hline
\end{tabular}}%
\caption{ Performance of \name\ on sentiment classification. Best scores are highlighted in bold.}
  \label{tab:results:sentiment}%
\end{table*}%
$\bullet$ \textbf{Sentiment classification:} We explore the Hinglish sentiment classification dataset developed by ~\newcite{joshi2016towards}. The dataset contains $3879$ Facebook public posts comprises of $15\%$ {\em negative}, $50\%$ {\em neutral}, and $35\%$ {\em positive} samples. We further consider two sentiment classification datasets for Dravidian languages \textit{viz.} Tamil and Malayalam \cite{chakravarthi-etal-2020-corpus}, 
containing $15744$ and $6739$ instances respectively with four sentiment labels -- {\em positive}, {\em negative}, {\em neutral}, and {\em mixed feelings}. Additionally, we use SemEval-2020 ~\cite{patwa-etal-2020-sentimix} dataset for Spanish code-mixed sentiment classification. It supports a classic 3-way sentiment classification.  \\
$\bullet$  \textbf{Named-entity recognition:} For NER, we employ Hindi \cite{singh2018named} and Spanish \cite{aguilar-etal-2018-named} datasets with $2079$ and $52781$ sentences, respectively. In Hindi, the labels are \textit{name}, \textit{location}, and \textit{organization}. The Spanish dataset has six additional labels -- \textit{event}, \textit{group}, \textit{product}, \textit{time}, \textit{title}, and \textit{other} named entities.\\
%
$\bullet$  \textbf{Machine Translation:} We utilize a recently developed Hindi-English code-mixed parallel corpus for machine translation \cite{gupta-etal-2020-semi} comprising more than $200k$ sentence pair. For experiments, we transliterate all Devanagari Hindi text.

\subsection{Baselines}
\label{subs:baseline}
\paragraph{POS tagging, NER \& sentiment classification:}

    \noindent $\rhd$ \textbf{BiLSTM} \cite{hochreiter1997long}: It is a weak baseline with two conventional BiLSTM layers. For POS and NER, we additionally incorporate a CRF layer for the final classification.
    $\rhd$ \textbf{HAN} \cite{yang2016hierarchical}: We adapt the Hierarchical Attention Network (HAN) for our purpose. The subword embedding is computed at the first level of attention network followed by a word-level attention at the second level. Recently, ~\newcite{bansal-etal-2020-code} also adopted HAN for code-mixed classification.
     $\rhd$ \textbf{ML-BERT} \cite{devlin-etal-2019-bert}: We fine-tune multilingual BERT \cite{m:bert}.
     $\rhd$ \textbf{CS-ELMo} \cite{aguilar-solorio-2020-english}: It is one of state-of-the-arts on code-mixed languages. It uses  pre-trained ELMo \cite{peters-etal-2018-deep} to transfer knowledge from English to code-mixed languages. 
    \noindent $\rhd$ \textbf{Subword-LSTM} \cite{joshi2016towards}: It is a hybrid CNN-LSTM model. A 1D convolution operation is applied for the subword representation. Subsquently, the convoluted features are max-pooled and fed to an LSTM. Since this system disregards word boundaries in a sentence, we use it for {\em sentiment classification only}.  
        

\paragraph{Machine translation:}
For machine translation, we evaluate {\name} against \textbf{GFF-Pointer} ~\cite{gupta-etal-2020-semi}, a gated feature fusion (GFF) based approach to amalgamate the XLM and syntactic features during encoding and a Pointer generator for decoding. Furthermore, we also incorporate three other baselines for comparison -- \textbf{Seq2Seq} ~\cite{sutskever2014advances}, \textbf{Attentive-Seq2Seq} ~\cite{bahdanau2014neural} and \textbf{Pointer Generator}~\cite{see2017get}.

\subsection{Experimental Setup}
For each experiment,
we use a $dropout=0.1$ in both transformer block and the task specific layers. Categorical cross-entropy loss with Adam ($\eta = 0.001$, $\beta_{1} = 0.9, \beta_{2} = 0.999$) optimizer \cite{kingma2014adam} is employed in all experiments. We train our models for maximum $500$ epochs with an early-stopping criteria having $patience = 50$. 
We additionally use a learning rate scheduler to reduce learning rate to $70\%$ at plateaus with a patience of $20$ epochs. All models are trained with \textit{batch-size}$=32$.



\begin{table*}[ht]
\centering
\resizebox{0.9\textwidth}{!}
{
\begin{tabular}{l|ccc|ccc|ccc||ccc}
\hline
 \multirow{2}{*}{\textbf{Model}} & \multicolumn{3}{c|}{\bf Hindi} & \multicolumn{3}{c|}{\bf Telugu} & \multicolumn{3}{c||}{\bf Bengali} & \multicolumn{3}{c}{\bf Spanish} \\
  &  \textbf{Pr.} & \textbf{Re.} & \textbf{F1} & \textbf{Pr.} & \textbf{Re.} & \textbf{F1} & \textbf{Pr.} & \textbf{Re.} & \textbf{F1} & \textbf{Pr.} & \textbf{Re.} & \textbf{F1} 
 \\
\hline
BLSTM-CRF & 0.821 & 0.913 & 0.782 & 0.595 & 0.747 & 0.572 & 0.842 & 0.851 & 0.817 & 0.704 & 0.836 & 0.680 \\
HAN & 0.802 & 0.879 & 0.815 & 0.693 & 0.701 & 0.684 & 0.811 & 0.823 & 0.818 & 0.497 & 0.629 & 0.527 \\
ML-BERT & 0.833 & 0.884 & 0.847 & 0.802 & 0.762 & 0.771 & 0.793 & 0.815 & 0.807 & 0.853 & 0.808 & 0.802 \\
CS-ELMO & 0.885 & \textbf{0.961} & 0.910 & 0.831 & 0.790 & 0.775 & \textbf{0.873} & 0.851 & 0.847 & 0.740 & \textbf{0.835} & 0.729 \\
\hline 
\rowcolor{blue!10} \textbf{\name} & \textbf{0.918} & 0.955 & \textbf{0.919} & 0.815 & 0.749 & 0.762 & 0.841 & \textbf{0.855} & \textbf{0.853} & \textbf{0.871} & 0.822 & \textbf{0.825}  \\ \hline 
(-) $Atn^{outer}$ & 0.893 & 0.948 & 0.914 & \textbf{0.839} & \textbf{0.793} & \textbf{0.786} & 0.839 & 0.852 & 0.845 & 0.859 & 0.813 & 0.820 \\ 
(-) \textit{char-level} \name & 0.686 & 0.922 & 0.708 & 0.629 & 0.758 & 0.626 & 0.802 & 0.830 & 0.819 & 0.723 & 0.796 & 0.732 \\ 
\hline


\end{tabular}}%
\caption{ Performance of \name\ on POS tagging. Best scores are highlighted in bold.}
  \label{tab:results:pos}%
\end{table*}%

\subsection{Experimental Results}
\label{subs:setup}
We compute precision, recall, F1-score for POS, NER, and sentiment classification, whereas, BLEU, METEOR, and ROUGE scores are reported for the machine translation task.

{\bf Sentiment classification:} As shown in Table \ref{tab:results:sentiment}, \name\ obtains best F1-scores across all languages. For Hindi, three baselines (BiLSTM, ML-BERT, and CS-ELMo) obtain the best F1-score of $0.909$, where \name\ yields a small improvement with $0.915$ F1-score. In comparison, we observe an improvement of $1.4\%$ for Tamil, where {\name} and the best baseline (CS-ELMo) report $0.473$ and $0.459$ F1-scores, respectively. We observe the same pattern for Malayalam and Spanish as well -- in both cases, \name\ obtains improvements of $0.9\%$ and $2.0\%$, respectively. For Malayalam, {\name} reports $0.651$ F1-score, whereas CS-ELMo reports $0.642$ F1-score. In case of Spanish, HAN turns out to be the best baseline with $0.440$ F1-score. 
Comparatively, \name\ achieves $0.460$ F1-score. 
The last two rows of Table \ref{tab:results:sentiment} report ablation results -- a) excluding outer-product attention ($Atn^{outer}$) from \name; and b) excluding sub-word embeddings (\textit{character-level} \name). In all cases, the absence of \textit{sub-word} embeddings yields negative effect on the performance; hence, suggesting the effectiveness of \textit{character-level} \name\ in the architecture. On the other hand, omitting outer-product attention declines F1-scores in $3$ out of $4$ cases -- we observe a margin improvement of $0.04$ points for Malayalam. In summary, \name\ attains state-of-the-art performance across all four datasets, whereas, the best baseline (CS-ELMo) reports $1.2\%$ lower scores on average. 

\begin{table}[t]
\centering
\resizebox{\columnwidth}{!}
{
\begin{tabular}{l|ccc||ccc}
\hline
 \multirow{2}{*}{\textbf{Model}} & \multicolumn{3}{c||}{\bf Hindi} & \multicolumn{3}{c}{\bf Spanish} \\
  &  \textbf{Pr.} & \textbf{Re.} & \textbf{F1} & \textbf{Pr.} & \textbf{Re.} & \textbf{F1} \\
\hline
BLSTM-CRF & 0.622 & 0.781 & 0.579 & 0.581 & 0.659 & 0.603 \\
HAN & 0.721 & 0.767 & 0.695 & 0.615 & 0.679 & 0.644 \\
ML-BERT & 0.792 & 0.779 & 0.714 & 0.652 & 0.623 & 0.643 \\
CS-ELMO & 0.815 & 0.780 & 0.735 & 0.683 & 0.668 & 0.671 \\
\hline
\rowcolor{blue!10} \textbf{\name} & \textbf{0.829} & 0.788 & \textbf{0.745} & \textbf{0.695} & \textbf{0.671} & \textbf{0.684} \\\hline
(-) $Atn^{outer}$ & 0.821 & 0.767 & 0.732 & 0.669 & 0.663 & 0.668 \\
(-) \textit{char-level} \name & 0.556 & \textbf{0.815} & 0.528 & 0.498 & 0.664 & 0.539 \\
\hline

\end{tabular}}%
\caption{ Performance of \name\ on NER.}
  \label{tab:results:ner}%
\end{table}%
{\bf POS tagging:} Table \ref{tab:results:pos} shows the comparative results for POS tagging in Hindi, Telugu, Bengali, and Spanish. Similar to sentiment classification, we observe that \name\ attains best F1-scores across three datasets ($0.625\%$ better on average). It achieves $0.919$, $0.762$, $0.853$, and $0.825$ F1-scores for Hindi, Telugu, Bengali, and Spanish, respectively. In comparison, CS-ELMo yields best F1-scores among all the baselines across three datasets \textit{viz.} Hindi ($0.910$), Telugu ($0.775$), and Bengali ($0.847$). For Spanish, ML-BERT obtains the best baseline F1-score of $0.802$.  From ablation, we observe the negative effect on performance by removing either the outer-product attention or \textit{character-level} \name\ for the majority of the cases.

{\bf NER:}
The performance of \name\ for NER is also in-line with the previous two tasks, as show in Table \ref{tab:results:ner}. As mentioned earlier, we evaluate \name\ for Hindi and Spanish datasets. In both cases, we observe $\ge1\%$ improvement in F1-score, in comparison with the best baseline (CS-ELMo).

In all three tasks, CS-ELMo is arguably the most consistent baseline. Together with the state-of-the-art performance of \name, we regard the good performance to the subword-level contextual modelling -- both systems use contextual representational models (ELMo and Transformer) to encode the syntactic and semantic features. Moreover, the FAME module in \name\ assists in improving the performance even further. 

\begin{table}
\centering
\scalebox{.75}
{
\begin{tabular}{l|c|c|c}
\hline
 \textbf{Model} & {\bf B} & {\bf R} & {\bf M} \\
\hline
Seq2Seq$\dagger$ & 15.49 & 35.29 & 23.72\\
Attentive-Seq2Seq$\dagger$ & 16.55 & 36.25 & 24.97 \\
Pointer Generator$\dagger$ & 17.62 & 37.32 & 25.61\\
GFF-Pointer$\dagger$ & 21.55 & 40.21 & 28.37 \\
Transformer & 21.83 & 42.19 & 27.89 \\
\hline
\rowcolor{blue!10} \textbf{\name} & \textbf{28.22} & \textbf{51.52} & \textbf{29.59} \\
\hline
(-) $Atn^{outer}$ & 25.95 & 49.19 & 27.63 \\ \hline
\end{tabular}}%
\caption{Performance of \name\ encoder-decoder on En-Hi Translation (\textbf{B}: BLEU, \textbf{R}: Rouge-L and \textbf{M}: METEOR). $\dagger$ Values are taken from \newcite{gupta-etal-2020-semi}.}
  \label{tab:results:NMT}%
\end{table}%

\begin{table*}
\centering
\subfloat[Hindi code-mixed\label{tab:transfer:hindi}]{
\resizebox{\columnwidth}{!}
{
\begin{tabular}{c|c|c||c|c|c}
\multicolumn{2}{c|}{} & \multirow{2}{*}{\textbf{Fine-tune}} & \multicolumn{3}{c}{\bf Target Tasks} \\ \cline{4-6}
\multicolumn{2}{c|}{} & & \textbf{PoS} & \textbf{NER} & \textbf{Sentiment} \\
\hline 

\hline
\multirow{6}{*}{\rotatebox{90}{\bf Source Tasks}}& \multirow{2}{*}{\bf PoS} & w/o & \multirow{2}{*}{0.919} & 0.702 & \textbf{0.890} \\
& & w/ & & 0.578 & 0.863 \\\cline{2-6}
& \multirow{2}{*}{\bf NER} & w/o & \textbf{0.924} & \multirow{2}{*}{0.745} & \textbf{0.885} \\
& & w/ & 0.873 & & \textbf{0.893} \\ \cline{2-6}
& \multirow{2}{*}{\bf Sentiment} & w/o & \textbf{0.936} & 0.729 & \multirow{2}{*}{0.871} \\
& & w/ & \textbf{0.928} & 0.691 & \\
\hline
\end{tabular}%
}
}
\subfloat[Spanish code-mixed\label{tab:transfer:spanish}]{
\resizebox{\columnwidth}{!}
{
\begin{tabular}{c|c|c||c|c|c}
\multicolumn{2}{c|}{} & \multirow{2}{*}{\textbf{Fine-tune}} & \multicolumn{3}{c}{\bf Target Tasks} \\ \cline{4-6}
\multicolumn{2}{c|}{} & & \textbf{PoS} & \textbf{NER} & \textbf{Sentiment} \\
\hline

\hline 
\multirow{6}{*}{\rotatebox{90}{\bf Source Tasks}}& \multirow{2}{*}{\bf PoS} & w/o & \multirow{2}{*}{0.825} & \textbf{0.710} & 0.417\\
& & w/ & & 0.656 & 0.419\\\cline{2-6}
& \multirow{2}{*}{\bf NER} & w/o & \textbf{0.881} & \multirow{2}{*}{0.648} & \textbf{0.473} \\
& & w/ & 0.663 & & \textbf{0.446} \\ \cline{2-6}
& \multirow{2}{*}{\bf Sentiment} & w/o & \textbf{0.918} & \textbf{0.969} & \multirow{2}{*}{0.445} \\
& & w/ & 0.732 & 0.687 & \\
\hline
\end{tabular}%
}
}

\caption{Transfer learning models. Code-mixed word representations, learned for a (source) task, is utilized for building models for other (target) tasks of same language w/ and w/o fine-tuning. We highlight the cases in bold where transfer learning achieves better performance than original base \name.}
\label{tab:transfer}%
\end{table*}%
{\bf Machine Translation:} Finally, Table \ref{tab:results:NMT} reports the results for the English to Hindi (En-Hi) machine translation task. For comparison, we also report BLEU, METEOR, and ROUGE-L scores for four baseline systems -- Seq2Seq \cite{sutskever2014advances}, Attentive-Se2Seq \cite{bahdanau2014neural}, Pointer Generator \cite{see2017get}, and GFF-Pointer \cite{gupta-etal-2020-semi}. For all three metrics, \name\ reports significant improvement ($1$-$9$ points) over the state-of-the-art and other baselines. GFF-Pointer obtains $21.55$ BLEU score, while the other baselines yield BLEU scores in the range $[15-17]$. In comparison, \name\ obtain $28.22$ BLEU, {\em an extremely convincing result}. Similarly, \name\ reports $51.52$ ROUGE  and $29.59$  METEOR scores, respectively.

\subsection{Effects of Transfer Learning across Tasks}
One of the core objectives of representation learning is that the learned representation should be task-invariant -- the representations learned for one task should also be (near) effective for other tasks. The intuition is that the syntactic and semantic features captured for a language should be independent of the tasks, and if it does not comply, the representation can be attributed to capture the task-specific feature, instead of linguistic features. To this end, we perform transfer learning experiments with (w/) and without (w/o) fine-tuning. Since we have only one dataset for Tamil, Telugu, Bengali, and Malayalam, we choose Hindi and Spanish code-mixed datasets (POS, NER, and sentiment classification) for the study. Table \ref{tab:transfer} reports results for both code-mixed languages. For each case, we learn \name's representation on one (source) task and subsequently utilize the representation for the other two tasks (targets). Moreover, we repeat each experiment with and without fine-tuning \name.   

\begin{table}[t]
    \centering
    \subfloat[Sentiment]{
    \resizebox{0.45\columnwidth}{!}
    {
    \begin{tabular}{c|c|c|c|}
         \multicolumn{1}{c}{} & \multicolumn{1}{c}{Pos} & \multicolumn{1}{c}{Neg} & \multicolumn{1}{c}{Neu} \\ \cline{2-4}
         Pos & \textcolor{blue}{\bf 0.85} & 0.05 & \bf 0.10 \\ \cline {2-4}
         Neg & 0.03 & \textcolor{blue}{\bf 0.87} & \bf 0.10 \\ \cline {2-4}
         Neu & - & 0.02 & \textcolor{blue}{\bf 0.98} \\ \cline {2-4}
    \end{tabular}}
    }
    
    \subfloat[NER]{
    \resizebox{\columnwidth}{!}{
    \begin{tabular}{l|c:c|c:c|c:c|c|}
         \multicolumn{1}{c}{} & \multicolumn{1}{c}{B-Per} & \multicolumn{1}{c}{I-Per} & \multicolumn{1}{c}{B-Loc} & \multicolumn{1}{c}{I-Loc} & \multicolumn{1}{c}{B-Org} & \multicolumn{1}{c}{I-Org} & \multicolumn{1}{c}{O} \\ \cline{2-8}
         B-Per & \textcolor{blue}{\bf 0.85} & 0.01 & - & - & - & - & \bf 0.14 \\ \cdashline {2-8}
         I-Per & 0.03 & \textcolor{blue}{\bf 0.87} & - & - & - & - & \bf 0.10 \\ \cline {2-8}
         B-Loc & - & - & \textcolor{blue}{\bf 0.91} & & 0.02 & 0.02 & 0.05 \\ \cdashline {2-8}
         I-Loc & - & 0.08 & & \textcolor{blue}{\bf 0.85} & - & - & 0.07 \\ \cline {2-8}
         B-Org & 0.02 & 0.02 & - & - & \textcolor{blue}{\bf 0.82} & - & \bf 0.14 \\ \cdashline {2-8}
         I-Org & - & - & - & - & - & \textcolor{blue}{\bf 0.71} & \bf 0.29 \\ \cline {2-8}
         O     & - & - & - & - & \bf 0.47 & - & \textcolor{blue}{\bf 0.53} \\ \cline {2-8}
         
    \end{tabular}}
    }
    
    \caption{Confusion matrices (in \%) for sentiment and NER on Hindi code-mixed dataset$^{\ref{lbl:confusion}}$.}
    \label{fig:confusion}
\end{table}

For Hindi code-mixed, we do not observe the positive effect of transfer learning for NER. It could be because of the limited lexical variations of named-entities in other datasets. However, we obtain the best F1-score ($0.936$) for POS tagging in a transfer learning setup with sentiment classification. Similarly, for the sentiment classification as target, we observe performance improvements with both POS and NER as source tasks. In Spanish, we also observe increment in F1-scores for all three tasks. We attribute these improvements to the availability of more number of sentences for \name\ to leverage the linguistic features in both Hindi and Spanish. 




\begin{table}[t]
    \centering
      \subfloat[Sentiment]{
    \resizebox{\columnwidth}{!}{
    \begin{tabular}{c|p{24em}|c|c|c}
         & \multirow{2}{*}{\bf Input} & \multirow{2}{*}{\bf Gold} & \multicolumn{2}{c}{\bf Prediction} \\ \cline{4-5}
         & & & \textbf{\name} & \bf CS-ELMo \\ \hline
         \multirow{2}{*}{1} & \textbf{Org:} \textit{safal videsh yatra ke liye badhai ho sir} & \multirow{2}{*}{Pos} & \multirow{2}{*}{Pos} & \multirow{2}{*}{\textcolor{red}{Neu}} \\ 
         & \textbf{Trans:} \textit{Congratulations on the successful foreign trip sir} & & & \\ \hline 
          \multirow{2}{*}{2} & \textbf{Org:} \textit{desh chodo pahaley yeh media ko change karo ... !! ?} & \multirow{2}{*}{Neu} & \multirow{2}{*}{\textcolor{red}{Neg}} & \multirow{2}{*}{\textcolor{red}{Neg}} \\  
          & \textbf{Trans:} \textit{Leave the country, first change the media} & & & \\ \hline 
    \end{tabular}}
    }   
    
    \subfloat[NER]{
      \resizebox{\columnwidth}{!}{
    \begin{tabular}{l|p{30em}}
        \hline
         \multirow{4}{*}{\rotatebox{90}{\bf Input}} & \textbf{Org:} \textit{@gurmeetramrahim \{\underline{dhan dhan satguru}\}$_{Per}$ tera hi aasra \#msgloveshumanity salute 2 \{\underline{msg}\}$_{Org}$ $<$url$>$} \\ 
         & \textbf{Translated:} \textit{@gurmeetramrahim we depend on you \{\underline{dhan dhan satguru}\}$_{Per}$ \#msgloveshumanity salute 2 \{\underline{msg}\}$_{Org}$ $<$url$>$} \\ \hline 
         \multirow{4}{*}{\rotatebox{90}{\bf Prediction}} & \textbf{\name:} \textit{@gurmeetramrahim \{\underline{dhan dhan satguru}\}$_{Per}$ tera hi aasra \#msgloveshumanity salute 2 \textcolor{red}{msg} $<$url$>$} \\
         & \textbf{CS-ELMo:} \textit{@gurmeetramrahim \{\underline{dhan dhan satguru}\}$_{Per}$ tera hi aasra \#msgloveshumanity salute 2 \{\underline{msg}\}$_{Org}$ $<$url$>$} \\ \hline
    \end{tabular}}
    }

    \caption{Error Analysis on Hindi code-mixed dataset.}
    \label{tab:error}
\end{table}

\begin{figure*}[t]
     \centering

    {\includegraphics[scale=.12]{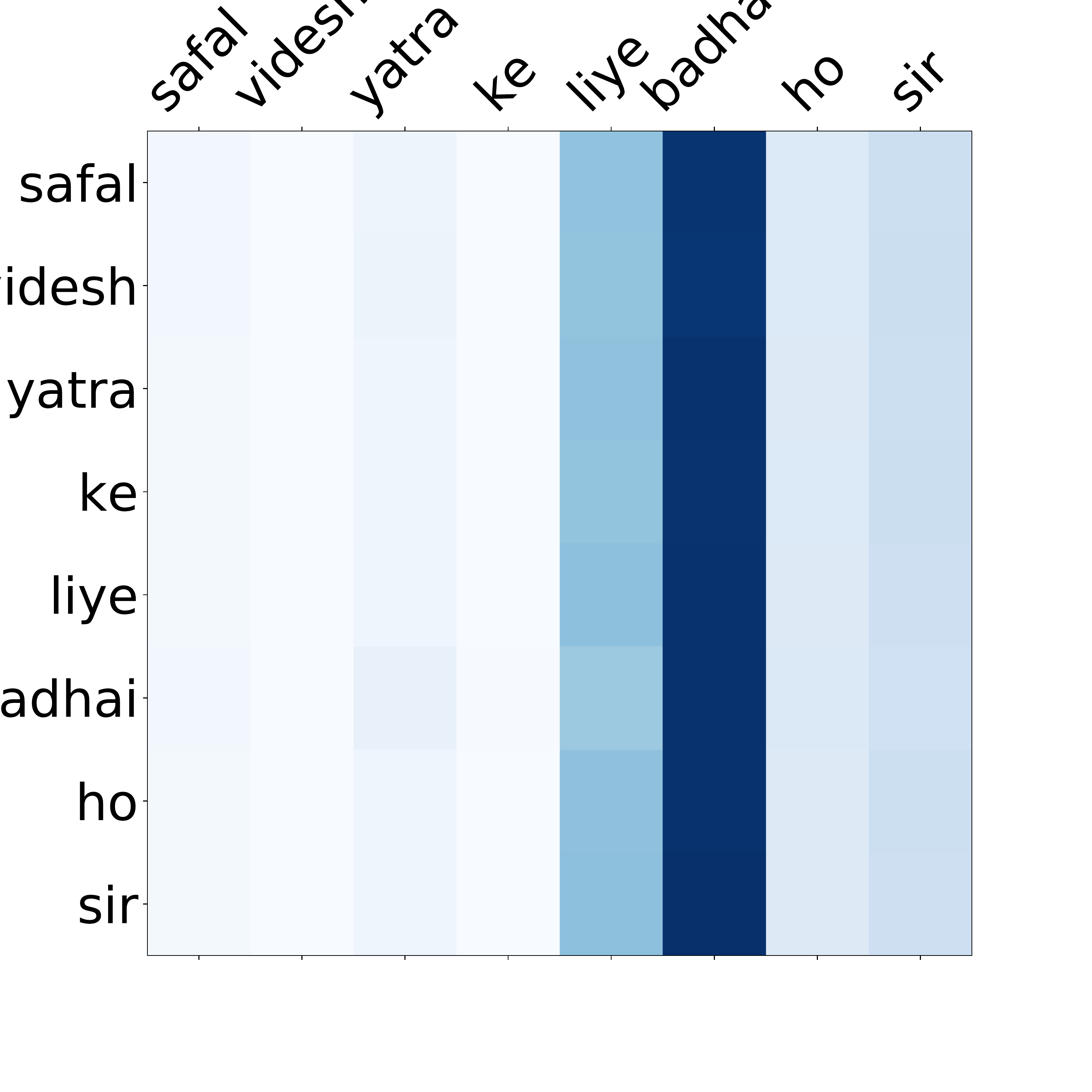}}
	 {\includegraphics[scale=.12]{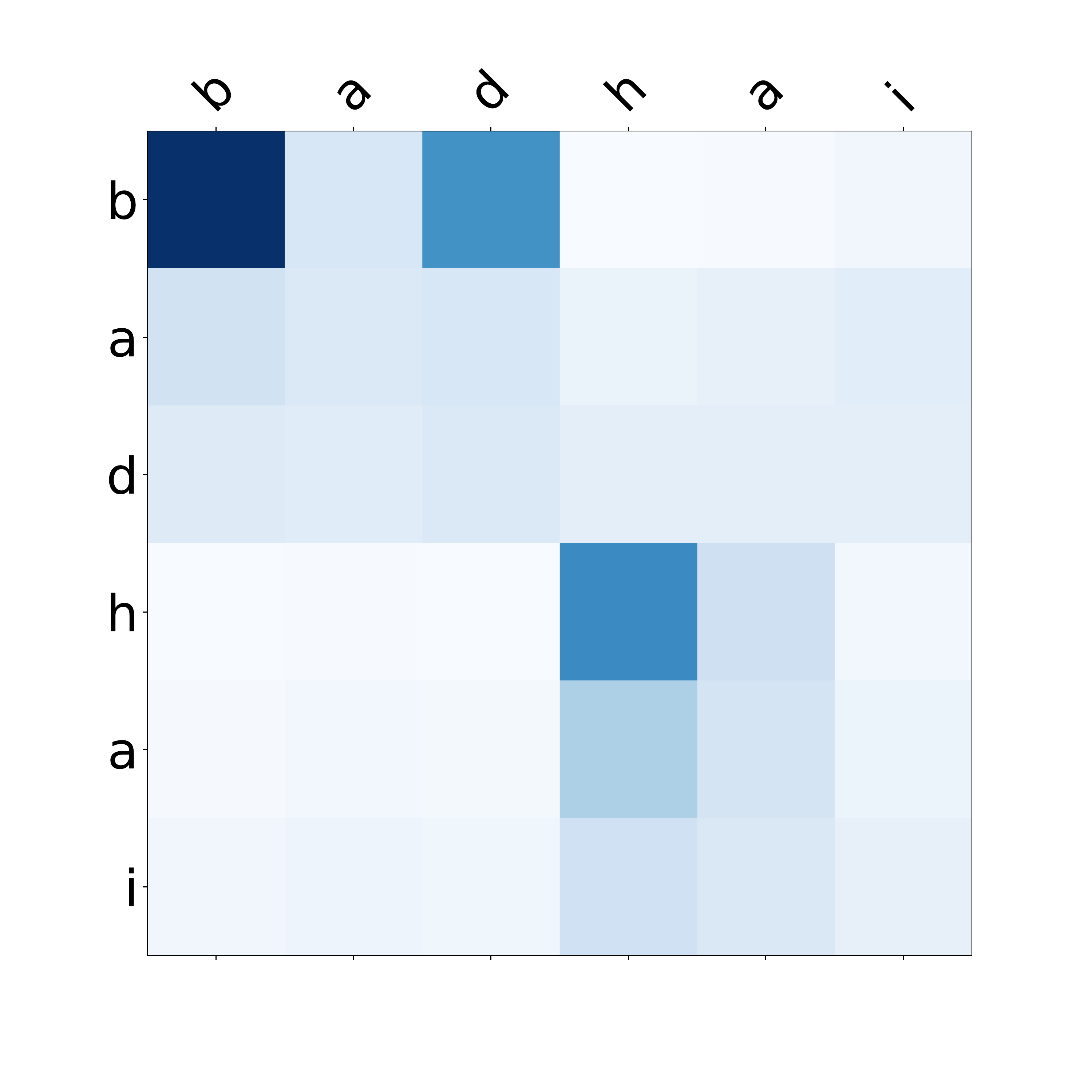}}
	 {\includegraphics[scale=.12]{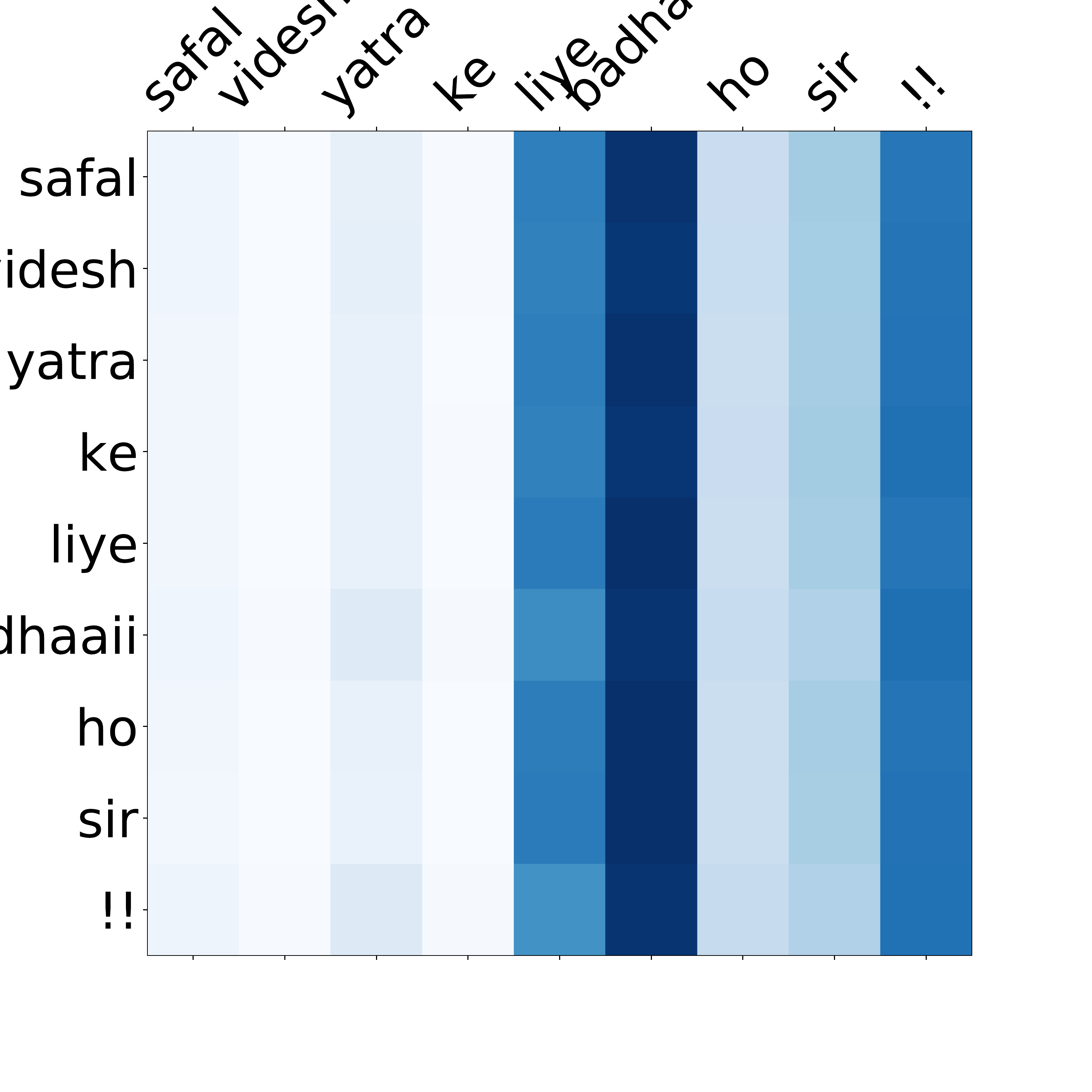}}
	 {\includegraphics[scale=.12]{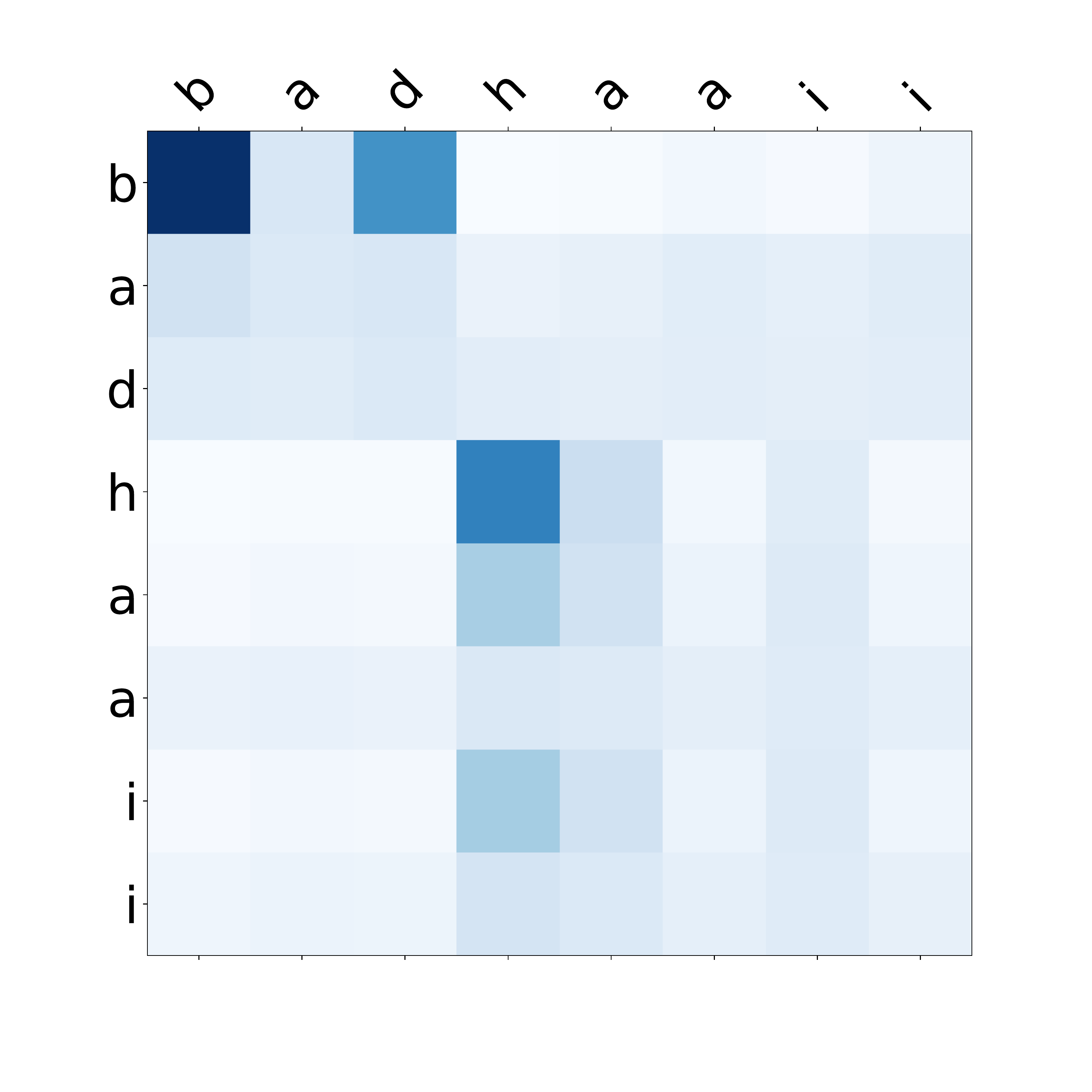}}

 	 \subfloat[Original input.\label{fig:interpretsentiment1:org}]{\includegraphics[scale=.5]{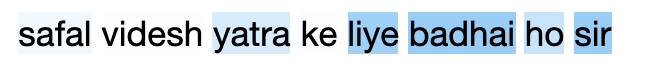}} \hspace{5em}
     \subfloat[Perturbed input.\label{fig:interpretsentiment1:perturb}]{\includegraphics[scale=.5]{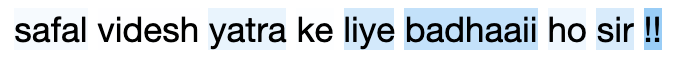}} \\
	 
	 \caption{Interpretation of Hindi code-mixed sentiment prediction (a) Grad-CAM  \cite{Selvaraju_2019} analysis of original text; (b) Grad-CAM of perturbed text (\textit{`badhaaii'}); For both case, the \textit{word-level} and \textit{char-level} attention plots are shown.
	 The impact of the word \textit{`badhai'} on the overall prediction is shown through gradients and attention heatmaps. The representation learning of phonetically similar word (e.g., \textit{`badhaaii'}) is also noticeable in the perturbed case. It signifies that HIT is flexible to the spelling variations, a common feature in code-mixed environment.}
     \label{fig:interpretsentiment1}
\end{figure*}

\subsection{Error Analysis}
\label{subs: analysis}
In this section, we analyze the performance of \name\ both quantitatively and qualitatively. At first, we report the confusion matrices\footnote{\label{lbl:confusion}Confusion matrices and more error cases for other tasks are presented in the appendix.} for Hindi NER and sentiment classification in Table \ref{fig:confusion}. In both cases, we observe the \textit{true-positives} to be significant for all labels. Furthermore, we also observe the \textit{false-positives} to be extremely low (except for `\textit{B-Org}' in NER) for majority of the cases -- suggesting very good precision in general. The major contribution in error is due to the \textit{neutral} and \textit{other} classes in sentiment and NER, respectively. In sentiment analysis, $10\%$ of the \textit{positive} and \textit{negative} labels each were mis-classified as \textit{neutral}. Similarly in NER, we observe that the \textit{organization} entities (\textit{B-Org \& I-Org}) and \textit{other} classes confuse with each other in a significant number of samples. It may be due to the under-represented ($\sim13\%$) \textit{organization} entities in the dataset.

We also perform qualitative error analysis  of \name\ and CS-ELMo. Table \ref{tab:error} reports the results for the NER and sentiment classification tasks$^{\ref{lbl:confusion}}$. For the first example in sentiment classification, \name\ accurately predicts the sentiment labels as \textit{positive}; however in comparison, CS-ELMo mis-classifies it as \textit{neutral}. For the second example, both \name\ and CS-ELMo wrongly predict the sentiment as \textit{neutral}. One possible reason could be the presence of the negatively-inclined word \textit{chodo} ({\em leave}) in the sentence. For NER, the sentence has two entities (one \textit{person} and one \textit{organization}). While \name\ correctly identifies `\textit{dhan dhan satguru}' as person, it could not recognize `\textit{msg}' as organization. On the other hand, CS-ELMo correctly identifies both.  


Furthermore, we take the first example of sentiment analysis (from Table \ref{tab:error}) to get the insight of \name.
It is not hard to understand that the most positive vibe is attributed by the phrase `\textit{badhai ho sir}' (\textit{congratulations sir}). 
To validate our hypothesis, we use a gradient based interpretation technique, Grad-CAM \cite{Selvaraju_2019}, which uses gradients of neural networks to show the effect of neurons on the final output. Due to hierarchical and modular nature of {\name}, we are able to extract the intermediate word level representations learnt by the \textit{character-level} \name\ and compute the gradient of loss of the actual class considering these representations. The magnitude of gradient shows the impact of each word on the final output. 
Figure \ref{fig:interpretsentiment1:org} shows the word-level and character-level gradient maps for the original input. We can observe that \name\ attends to the most important component in both cases. For \textit{word-level}, it highlights the positive phrase `\textit{liye badhai}' ({\em congratulations on}). Moreover, character-level \name\ attends to the two syllables `\textit{b}' and `\textit{dh}' in the word `\textit{badhai}' ({\em congratulation}). It suggests that both the \textit{word-level} and \textit{character-level} components are capable of extracting important features from inputs. Furthermore, to check the robustness, we investigate \name\ on a perturbed input. In the previous example, we tweak the spelling of the most important word `\textit{badhai}' to `\textit{badha\textbf{\underline{a}}i\textbf{\underline{i}}}' (an out-of-vocabulary word considering the dataset). Figure \ref{fig:interpretsentiment1:perturb} shows similar patterns in the perturbed input case as well. It signifies that \name\ identifies the phonetic similarity of the two words and is flexible to the spelling variants, a common feature in code-mixed environment.

\subsection{\name's Performance on Monolingual Data} \label{sec:monolingual}
In this section, we outline the performance of {\name} for monolingual and low-resource settings. We consider the sentiment classification dataset curated by \newcite{akhtar-etal-2016-aspect}, containing 5417 transliterated \emph{Hindi} reviews with 4 sentiment labels - \emph{positive}, \emph{negative}, \emph{neutral}, and \emph{conflict}. We also utilize a Magahi POS dataset \cite{kumar-etal-2012-developing}, annotated with $33$ tags from the BIS-tagset \footnote{\url{https://thottingal.in/blog/2019/09/10/bis-pos-tagset-review/}}. We report the performance of \name\ and other baselines on these two datasets in Table \ref{tab:monolingual}. For the Hindi sentiment classification task, we observe that {\name} yields an F1-score of $0.635$, which is better than CS-ELMo and ML-BERT by $9.3\%$ and $5.9\%$. Also, for Magahi POS, {\name} reports the best F1-score of $0.775$ -- increaments of $+2.1\%$ and $+9.5\%$ over CS-ELMo and ML-BERT, respectively. These results suggest that {\name} is capable of handling monolingual and low-resource texts in an efficient manner.

\begin{table}[t]
\centering
\resizebox{.9\columnwidth}{!}
{
\begin{tabular}{l|ccc|ccc}
\hline
 \multirow{2}{*}{\textbf{Model}} & \multicolumn{3}{c|}{\bf Hi Sentiment}  & \multicolumn{3}{c}{\bf Magahi POS} \\
 \cline{2-7} 
  &  \textbf{Pr.} & \textbf{Re.} & \textbf{F1} & \textbf{Pr.} & \textbf{Re.} & \textbf{F1} \\
\hline
BiLSTM & 0.619 & 0.533 & 0.554 & 0.594 & 0.804 & 0.626 \\
HAN & 0.602 & 0.528 & 0.551 & 0.729 & 0.857 & 0.649 \\
ML-BERT & 0.604 & 0.556 & 0.576 & 0.757 & 0.867 & 0.708 \\
CS-ELMO & 0.593 & 0.520 & 0.542 & 0.771 & 0.884 & 0.759\\
\hline 
\rowcolor{blue!10} \textbf{\name} & \textbf{0.641} & \textbf{0.629} & \textbf{0.635} & \textbf{0.783} & \textbf{0.913} & \textbf{0.775} \\ \hline 

\end{tabular}}%
\caption{ Performance of {\name} on monolingual tasks. Best scores are highlighted in bold.}
  \label{tab:monolingual}%
\end{table}%

\subsection{Learnable Parameters and Power Usage}
We conduct all our experiments on 1 Tesla T4 GPU. In Table \ref{tab:parameters}, we report the total trainable parameters for HIT and other baselines. We observe that \name\ requires a comparable number of parameters. For instance, in the Hindi-English sentiment analysis task (sequence classification), HIT has a total \textasciitilde$2.7M$ trainable parameters, while other baselines such as, CS-ELMo, HAN, Subword-LSTM, and BiLSTM require \textasciitilde$2.9M$, \textasciitilde$2.7M$, \textasciitilde$2.1M$, and \textasciitilde$2.8M$ parameters, respectively. ML-BERT has a whopping \textasciitilde$179.2M$ parameters. Similarly, in Hindi-English POS tagging, the number of parameters for HIT is comparable (or even lesser) -- HIT: \textasciitilde$1.4M$, CS-ELMo: \textasciitilde$2.4M$, HAN: \textasciitilde$1.4M$, BiLSTM-CRF: \textasciitilde$1.5M$, ML-BERT: \textasciitilde$177.9M$. We observed similar distribution for other tasks/languages as well.

We further note that {\name} is significantly more efficient than the current SOTA models as it takes $13$ s/epoch to train which is significantly lower than CS-ELMo ($18$ s/epoch), HAN ($14$ s/epoch), and ML-BERT ($172$ s/epoch), while it takes a bit more time compared to BiLSTM ($12$ s/epoch) and Subword-LSTM ($7$ s/epoch). We also computed the amount of power consumption for training \name\ for a maximum 500 epochs. Following the guidelines of \newcite{strubell-etal-2019-energy}, we estimate a total power consumption of 0.383 kWh and equivalent CO2 emission of 0.365 pounds. 

\begin{table}[t]
    \centering
      \resizebox{.9\columnwidth}{!}{
        \begin{tabular}{c|l|c|c}
         {\bf Tasks} & {\bf Model} & {\bf \# Params(M)} & {\bf Train Time(s/ep)}\\
          \hline
         \multirow{5}{*}{\rotatebox{90}{\bf Sentiment}} & CS-ELMo & 2.9 & 18 \\
          & HAN & 2.7 & 14\\
          & BiLSTM & 2.8 & 12 \\
          & ML-BERT & 179.2 & 172 \\
          & \cellcolor{blue!10} {\bf \name} & \cellcolor{blue!10}2.7 & \cellcolor{blue!10}13  \\
          \hline
          \multirow{5}{*}{\rotatebox{90}{\bf POS}} & CS-ELMo & 2.4 & 4\\
          & HAN & 1.4 & 3\\
          & BiLSTM-CRF & 1.5 & 2\\
          & ML-BERT & 177.9 & 114\\
          & \cellcolor{blue!10} {\bf \name} & \cellcolor{blue!10}1.4 & \cellcolor{blue!10}2\\
          \hline
    \end{tabular}}
    \caption{Parameters and Runtime: Number of trainable parameters and training runtime (second/epoch) for the Hi-En PoS and sentiment classification tasks.}
    \label{tab:parameters}
\end{table}

\section{Conclusion}
In this work, we present {\name} -- a hierarchical transformer-based framework for learning robust code-mixed representations. \name\ contains a novel fused attention mechanism, which calculates a weighted sum of the multi-headed self attention and outer-product attention, and is capable of capturing relevant information at a more granular level. We experimented with eleven code-mixed datasets for POS, NER, sentiment classification, and MT tasks across six languages. We observed that {\name} successfully outperforms existing SOTA systems. We also demonstrate the task-invariant nature of the representations learned by {\name} via a transfer learning setup, signifying it's effectiveness in learning linguistic features of CM text rather than task-specific features. Finally, we qualitatively show that {\name} successfully embeds semantically and phonetically similar words of a code-mixed language. 

\section*{Acknowledgement}
The work was partially supported by the Ramanujan Fellowship (SERB) and the Infosys Centre for AI, IIITD.

\bibliographystyle{acl_natbib}
\bibliography{acl2021}

\newpage
\appendix

\section{Appendix}
\subsection{Semantic Understanding of Languages}\label{subs: semanticanalysis}

In this section, we study the semantic relationships between different Indic languages. We calculate the proportion of common words in Table~\ref{tab:words} between different language pairs to understand the multilingualism in India. We observe that Bengali code-mixed texts have the highest proportion of English words $32\%$ as compared to other languages. Moreover, $50\%$ of all Bengali words are also present in the Hindi CM texts, although $58\%$ of those words are English. We observe that users using Hindi CM texts use very few words taken from other languages. On the other hand, a significant proportion of Bengali and Telugu CM words are common in other languages, although, majority of them are English. The two Dravidian languages, Tamil and Malayalam, show a very distinctive behavior. They share very little linguistic similarity with other Indic languages. On the other hand, $10\%$ of all Tamil words are used in Malayalam and $17\%$ of all Malayalam words are used in Tamil. Moreover, this sharing is not driven by English, as, only $27\%$ of these words are English, which is the lowest proportion among all other language pairs. Being originate from a similar root and having a phonetic resemblance makes Tamil and Malayalam \textit{sister languages}\footnote{\url{https://royalsocietypublishing.org/doi/10.1098/rsos.171504}}.
\begin{figure}[ht!]
     \centering
     \subfloat[]{\includegraphics[scale=.2]{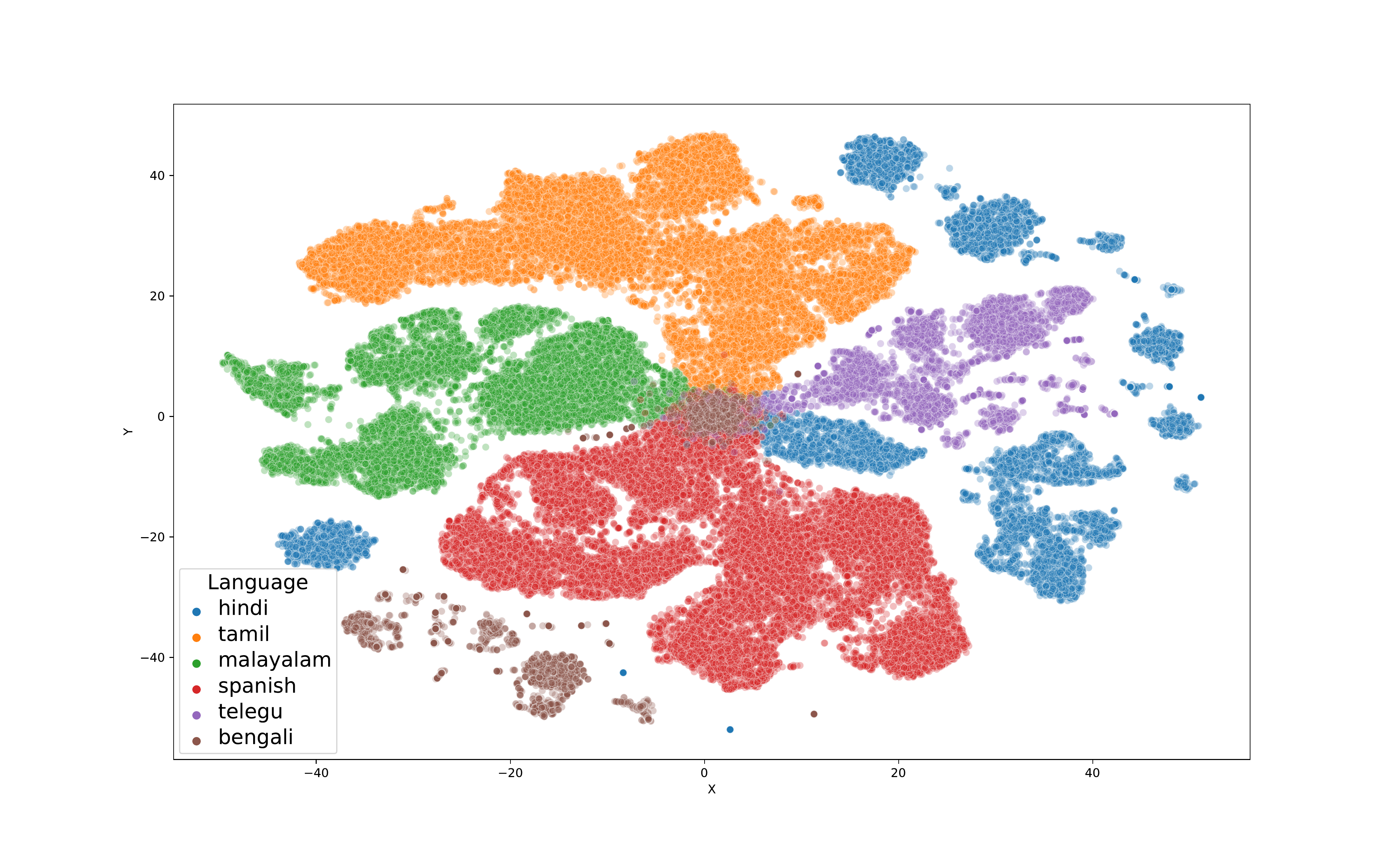}}
     
     \subfloat[]{\includegraphics[scale=.2]{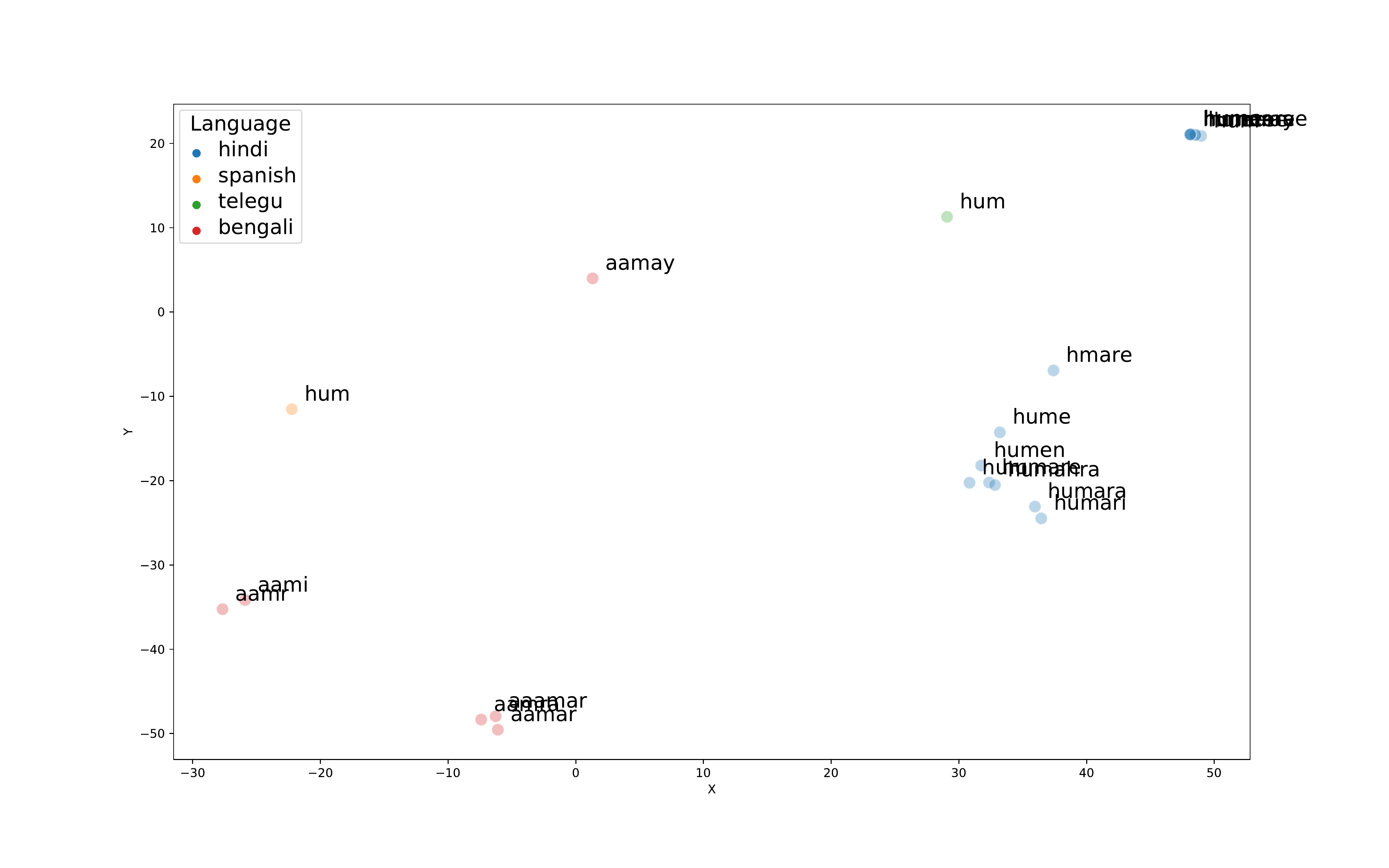}}
     \caption{t-SNE visualization (a) of all words;\, (b) of selected pronouns. Overlapping clusters show how semantically similar words from different languages are embedded onto a similar space.}
     \label{fig:tsne}
\end{figure}
\begin{table*}[ht!]
\centering
\scalebox{0.6}
{
\begin{tabular}{c|l|c|c|c|c|c|c}
\multicolumn{2}{c|}{} & \multicolumn{6}{|c}{\bf Target Language} \\ \cline{3-8}
\multicolumn{2}{c|}{} & \textbf{Hindi (English)} & \textbf{Malayalam (English)} & \textbf{Tamil (English)} & \textbf{Bengali (English)} & \textbf{Telegu (English)} & \textbf{Spanish (English)}\\
\hline 
\multirow{6}{*}{\rotatebox{90}{\bf Source Language}} & Hindi & 1.00 (0.16) & 0.02 (0.41) & 0.04 (0.39) & 0.02 (0.58) & 0.02 (0.57) & 0.07 (0.62) \\\cline{2-8}
& Malayalam & 0.14 (0.41) & 1.00 (0.06) & 0.17 (0.27) & 0.03 (0.71) & 0.05 (0.57) & 0.07 (0.64) \\\cline{2-8}
& Tamil & 0.15 (0.39) & 0.10 (0.27) & 1.00 (0.07) & 0.03 (0.69) & 0.05 (0.56) & 0.07 (0.64) \\\cline{2-8}
& Bengali & 0.50 (0.58) & 0.16 (0.58) & 0.23 (0.69) & 1.00 (0.32) & 0.21 (0.71) & 0.36 (0.72) \\\cline{2-8}
& Telegu & 0.36 (0.57) & 0.15 (0.57) & 0.29 (0.56) & 0.12 (0.71) & 1.00 (0.22) & 0.28 (0.65) \\\cline{2-8}
& Spanish & 0.12 (0.62) & 0.02 (0.64) & 0.03 (0.64) & 0.02 (0.72) & 0.03 (0.65) & 1.00 (0.11) \\\cline{2-8}
\hline
\end{tabular}%
}
\caption{Proportion of words in source language in the target language.}
\label{tab:words}
\end{table*}

\begin{table*}[ht!]
    \centering
    \resizebox{\textwidth}{!}{
    \begin{tabular}{c|p{40em}}
         \textbf{Lang} & \textbf{POS tags} \\ \hline
         Hindi (14) & \textit{X, VERB, NOUN, ADP, PROPN, ADJ, PART, PRON, DET, ADV, CONJ, PART\_NEG, PRON\_WH, NUM}\\ \hline
         Bengali (39) & \textit{N\_NN, V\_VM, RD\_PUNC, N\_NNP, PSP, PR\_PRP, JJ, RB\_AMN, CC, QT\_QTF, DM\_DMD, RP\_RPD, @, RD\_RDF, V\_VAUX, DT, PR\_PRQ, \#, RP\_NEG, E, \$, RB\_ALC, N\_NNV, PR\_PRL, N\_NST, RP\_INJ, RD\_SYM, DM\_DMR, RP\_INTF, PR\_PRF, DM\_DMQ, QT\_QTO, U, QT\_QTC, PR\_PRC, RD\_ECH, QY\_QTO, Ã°Å¸Ëœ, $\sim$} \\ \hline
         Telugu (52) & \textit{N\_NN, N\_NNP, RD\_RDF, RD\_PUNC, V\_VM, JJ, @, PSP, PR\_PRP, RP\_INJ, DT, RB\_AMN, CC, \$, U, E, \#, N\_NNV, \&, PR\_PRQ, V\_VAUX, RD\_PUNC", $\sim$, RD\_RDFP, QT\_QTF, RD\_UNK, DM\_DMD, RP\_RPD, RB\_ALC, DM\_DMQ, RD\_ECH, N\_NST, acro, PR\_PRL, QT\_QFC, RP\_RDF, PR\_PRC, r, RD\_SYM, RD\_RDFF, psp, PR\_PRF, QT\_QTP, RD\_P/UNC, PR\_PPR, PR\_RPQ, RPR\_PRP, RP\_INTF, -} \\ \hline
         Spanish (17) & \textit{VERB, PUNCT, PRON, NOUN, DET, ADV, ADP, INTJ, CONJ, ADJ, AUX, SCONJ, PART, PROPN, NUM, UNK, X}\\ \hline
    \end{tabular}}
    \caption{POS tagsets for different datasets.}
    \label{tab:tags}
\end{table*}
Similar observations are also made from the word representation lens. We use t-SNE~\cite{van2008visualizing} plots to embed {\name's} representations onto a $2$-D space for interpretability (Fig~\ref{fig:tsne}). Although, the embeddings are well clustered based on the languages, we can easily figure out the semantically similar words across languages embedded onto a similar space. Furthermore, Fig~\ref{fig:tsne}(b) shows that pronouns (e.g., \textit{`aap'}) in Tamil, Telegu and Hindi are embedded onto a similar space with Bengali words \textit{`aamar', `aamay'}. Although each of these representations are learned on separate models on separate datasets, the robustness of the underlying hierarchical representation enables our model to capture cross-lingual semantics from noisy code-mixed texts. We can attribute these observations to the relatedness of Indic languages on a socio-cultural basis.

\subsection{Datasets}
We report all available POS tags in Table \ref{tab:tags}.\\

\subsection{Confusion Matrices and Error Analysis}
\begin{figure*}[ht!]
    \centering
    \subfloat[Hindi]{\includegraphics[scale=.17]{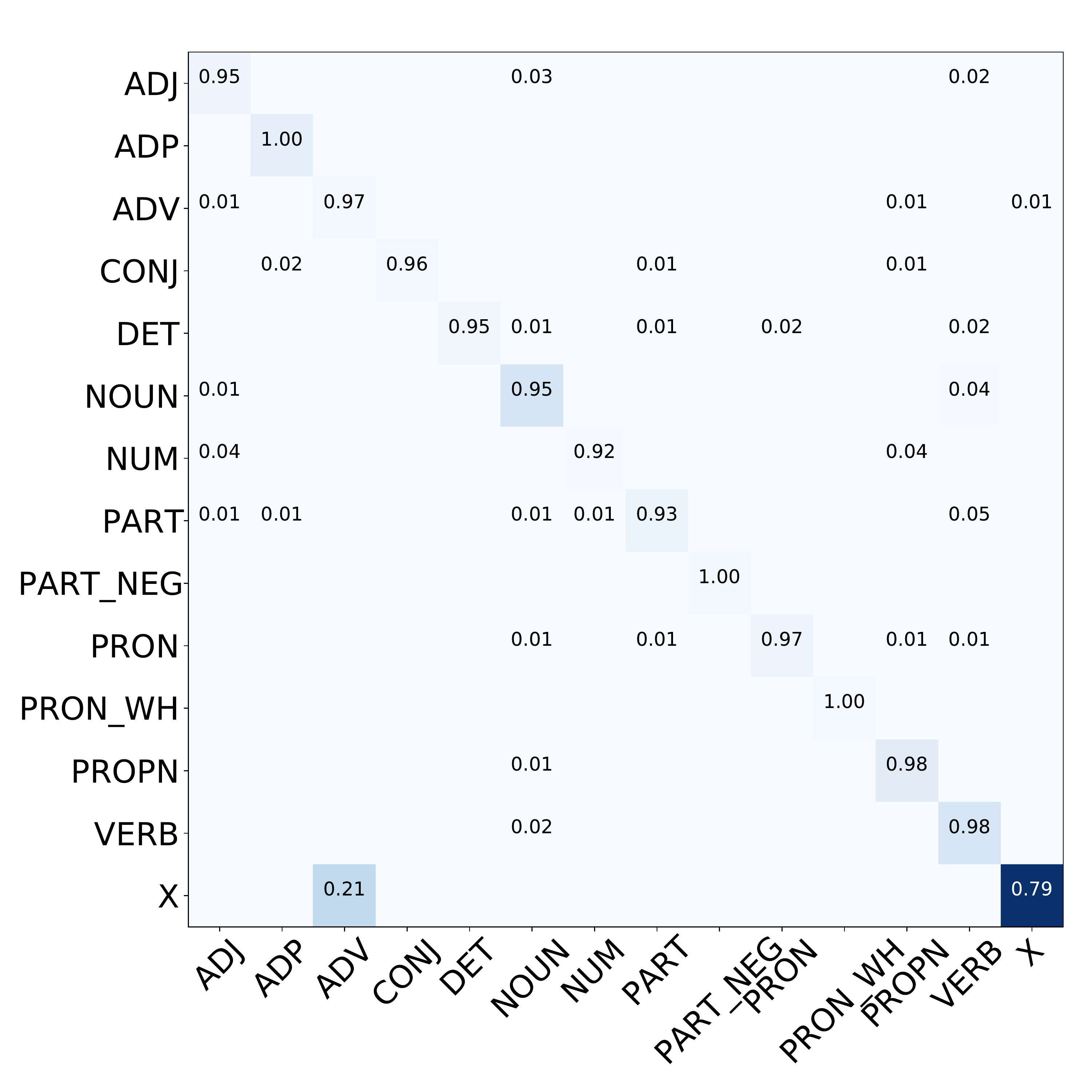}} 
    \subfloat[Spanish]{\includegraphics[scale=.17]{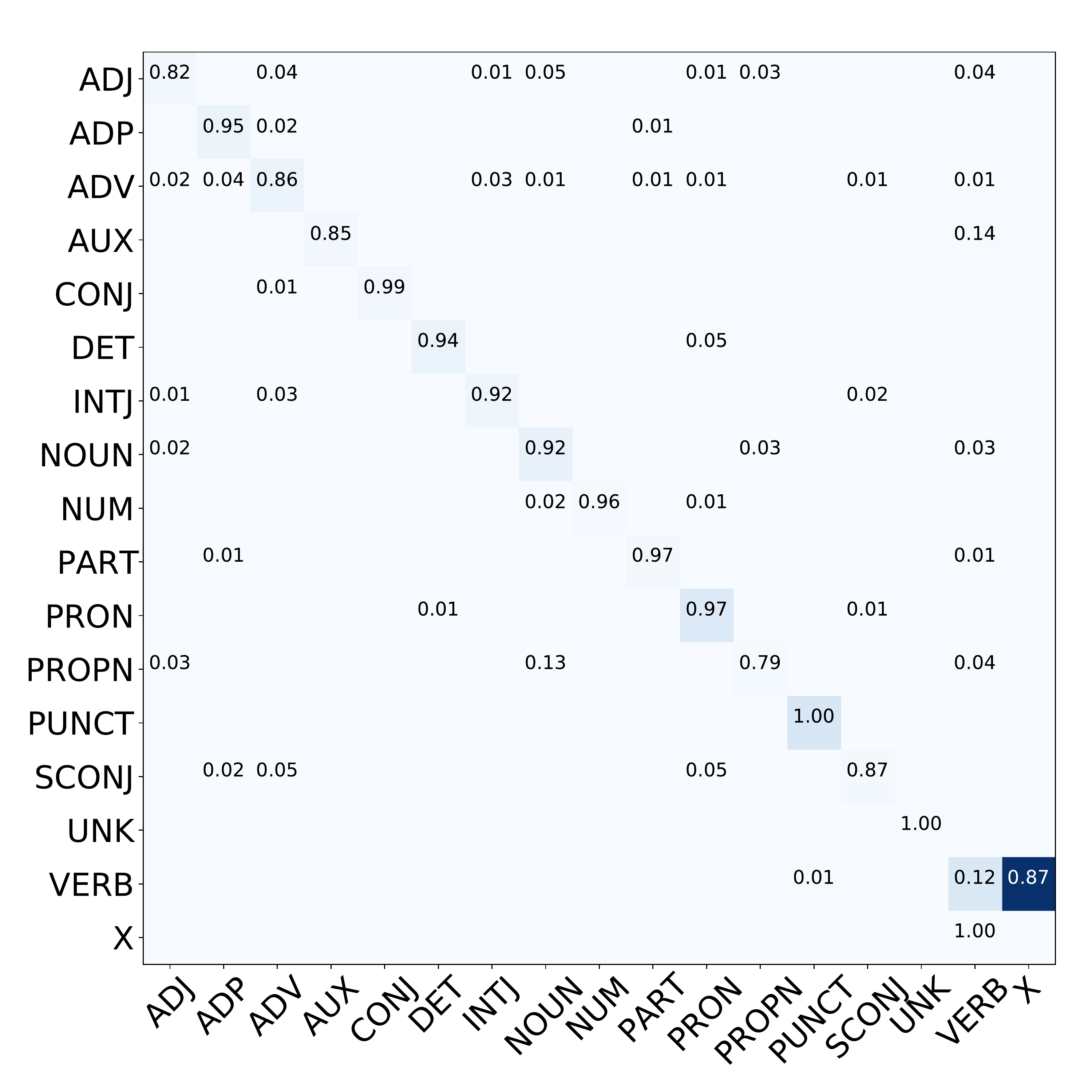}}
    \caption{Confusion matrices of {\tt HIT} on POS tasks. Due to high cardinality of output classes, we do not report for Bengali and Telugu.}
    \label{fig:posconfusion}
\end{figure*}

\begin{figure*}[ht!]
     \centering
     \subfloat[Tamil]{\includegraphics[scale=.17]{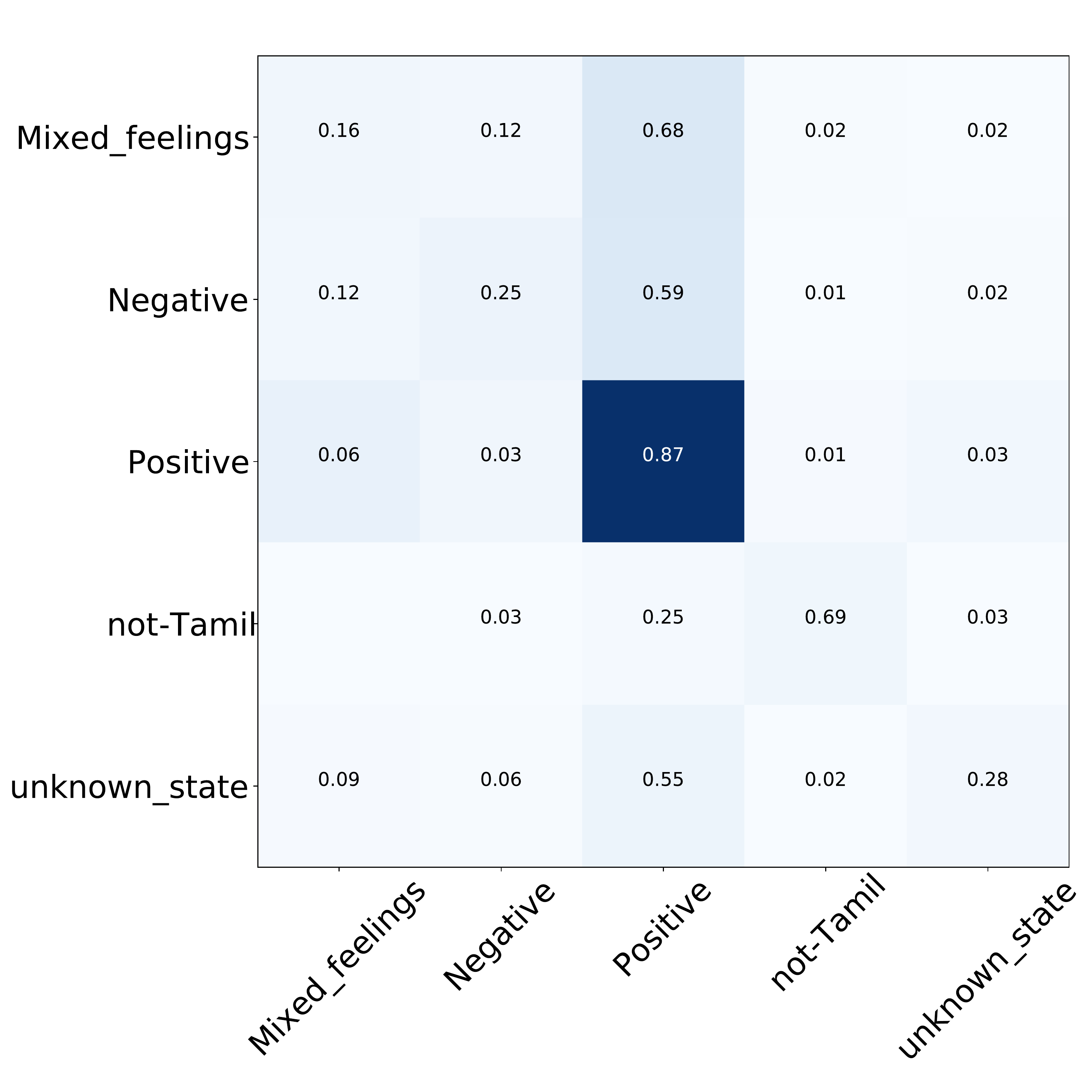}}
	 \subfloat[Malayalam]{\includegraphics[scale=.17]{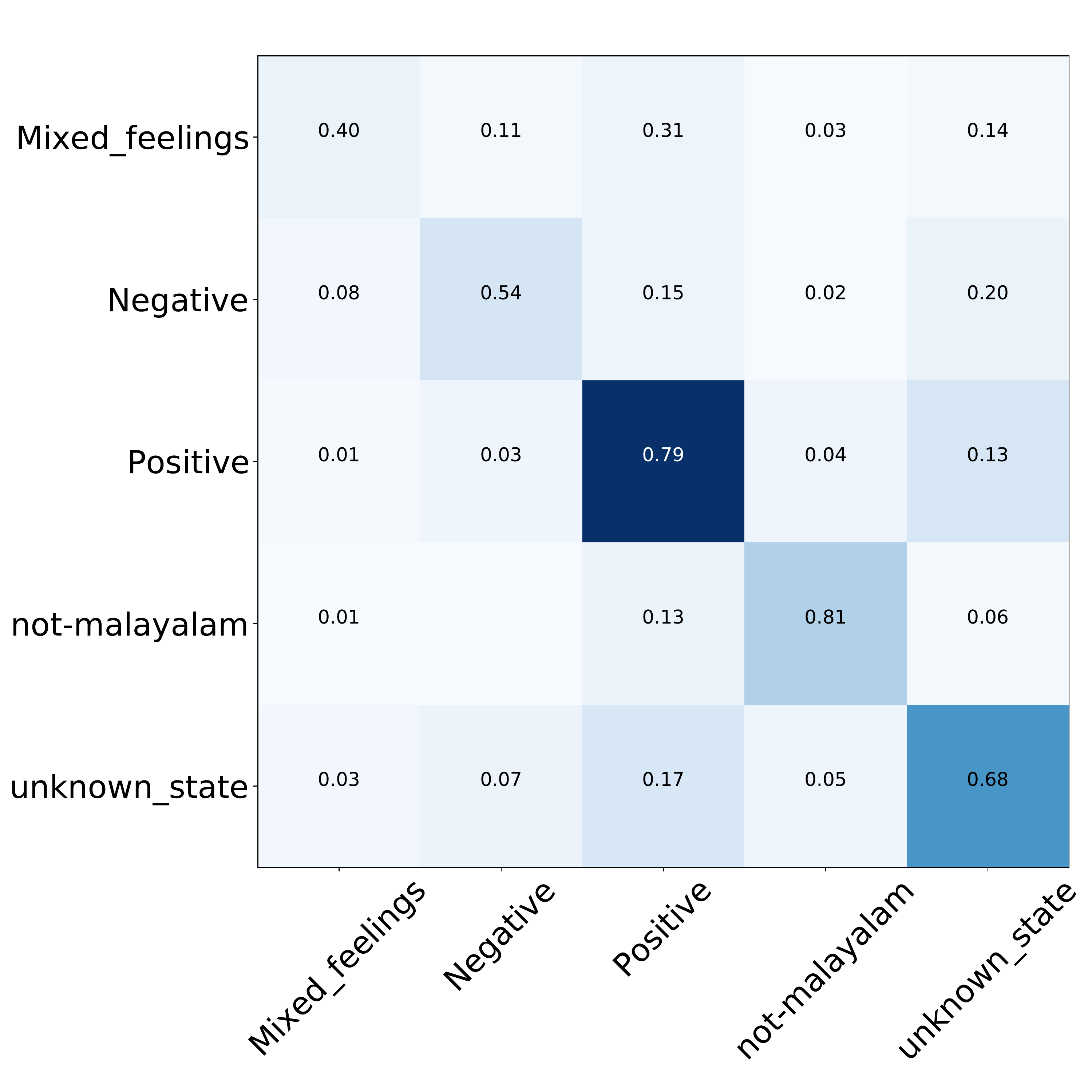}}
	 \subfloat[Spanish]{\includegraphics[scale=.17]{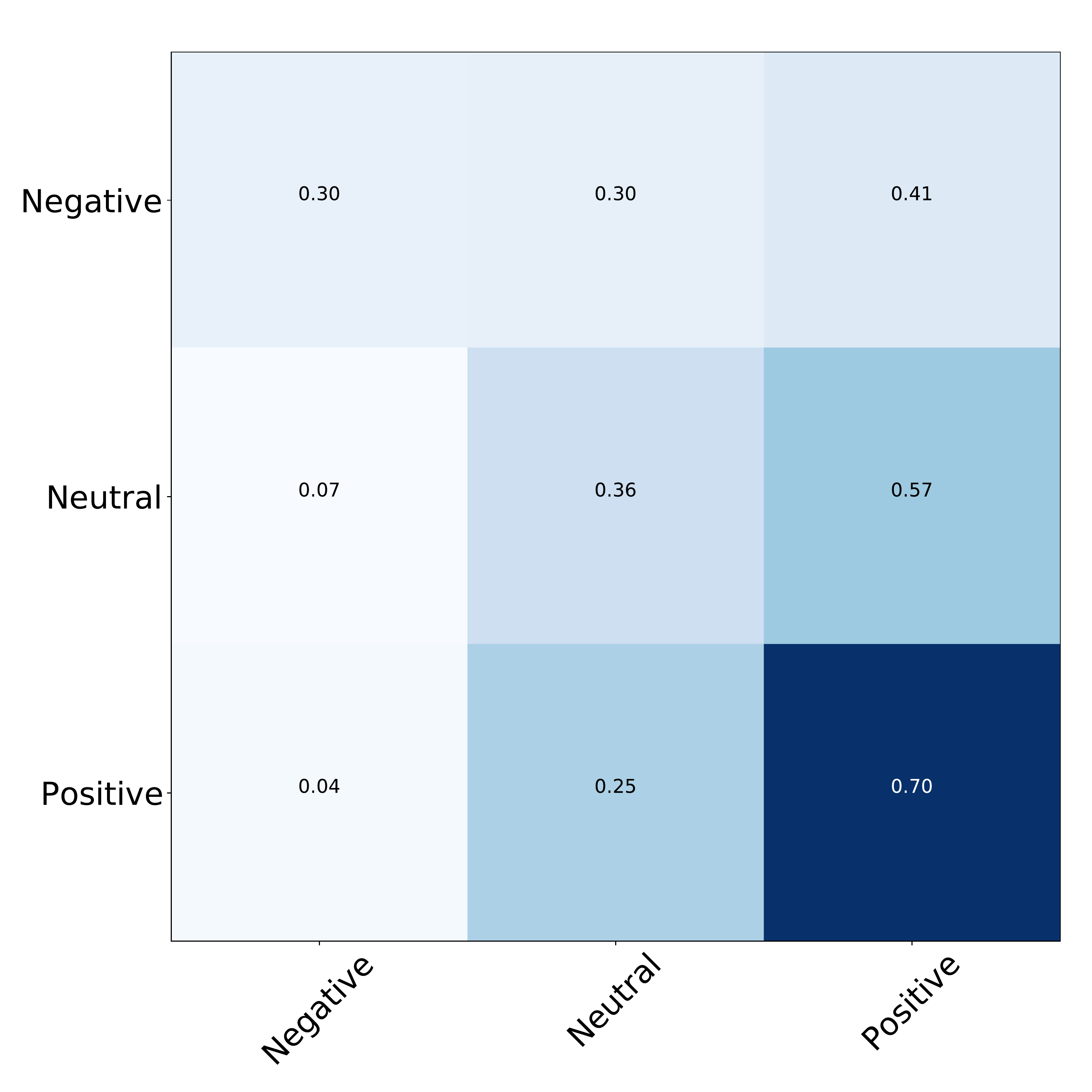}}
     \caption{Confusion matrices of {\tt HIT} on sentiment tasks.}
     \label{fig:senticonfusion}
\end{figure*}

\begin{figure*}[ht!]
    \centering
     \subfloat[Hindi]{\includegraphics[scale=.1724]{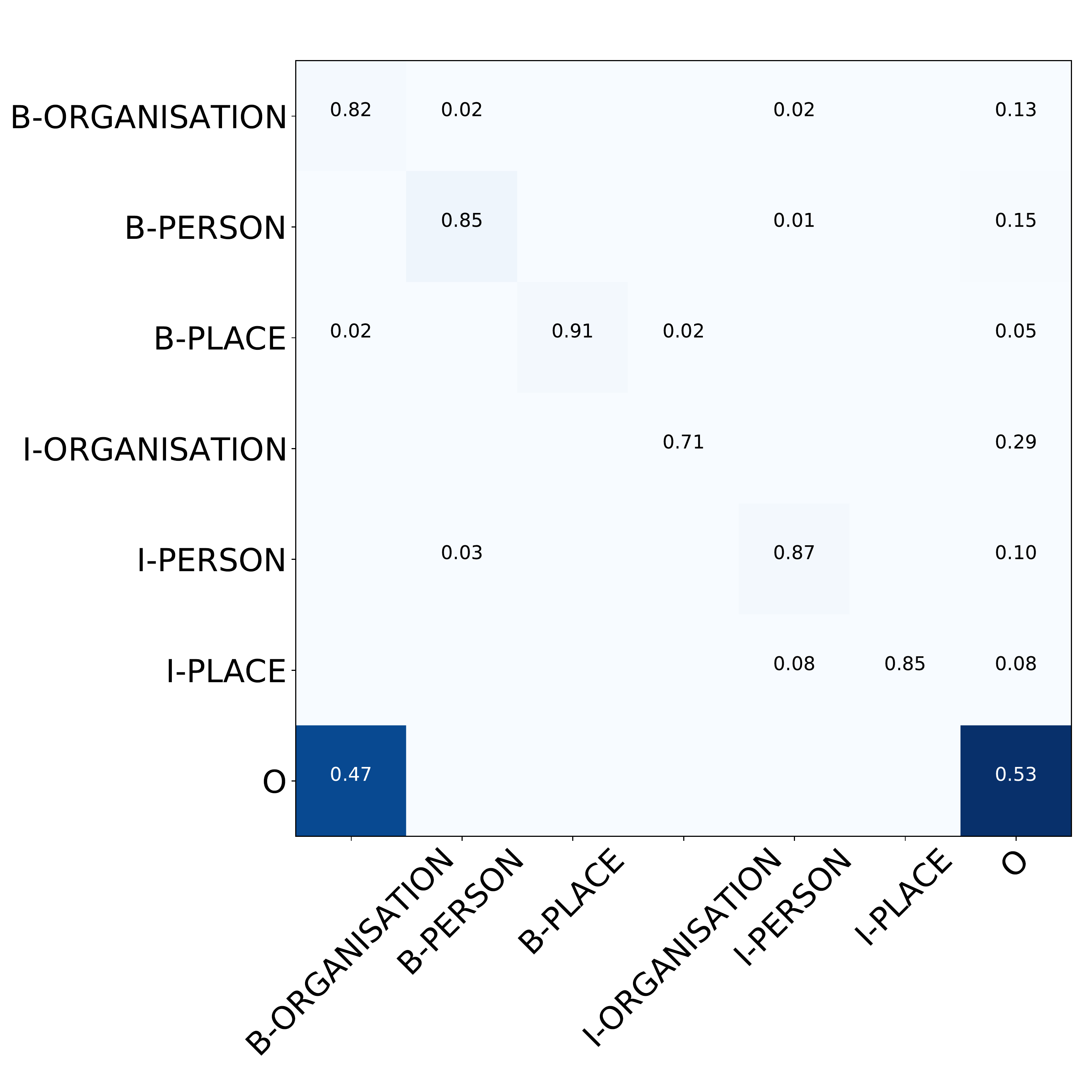}} \hspace{2em}
    \subfloat[Spanish]{\includegraphics[scale=.1724]{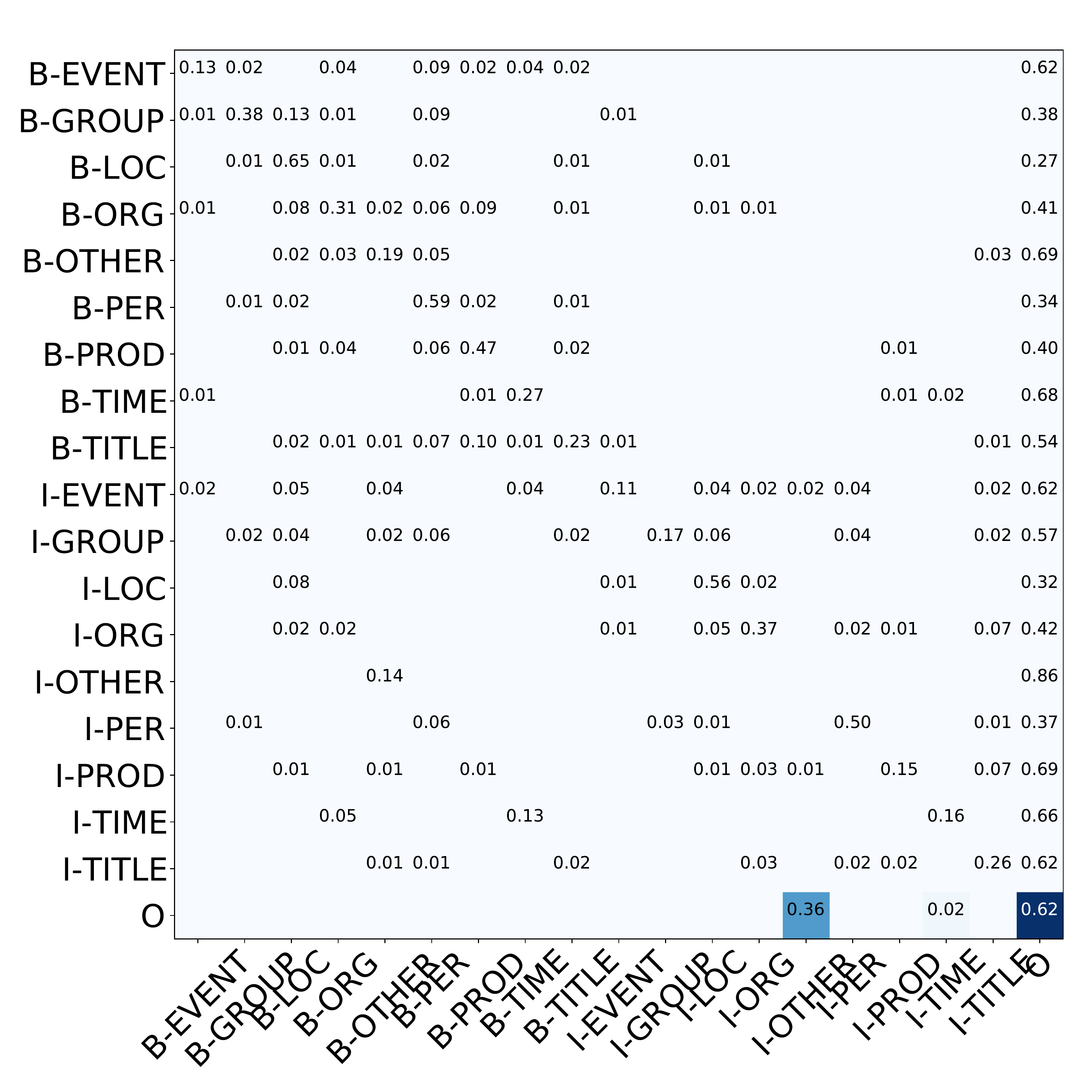}}
    \caption{Confusion matrices of {\tt HIT} on NER.}
    \label{fig:nerconfusion}
\end{figure*}

We report the confusion matrices to show the label-wise performance for the sentiment classification, PoS tagging and NER in Tables \ref{fig:senticonfusion}, \ref{fig:posconfusion}, and \ref{fig:nerconfusion}, respectively. 

We similarly perform qualitative analysis on the MT task where our model shows superior performance as compared to the baselines. In example 1 of Table~\ref{tab:error:supp} (d), {\name} translates the English text `\textit{`Licencing and import policies were liberalise''} to \textit{``Licencing aur policies liberal the |''}. Although this prediction has very low BLEU score when evaluated against the target, this example shows an interesting observation. 
The overall translation is a contextually meaningful sentence in Hindi. Further {\name} translates the phrase \textit{`were liberalise'} to \textit{`liberal the'}. In Hindi, \textit{`the'} represents past tense. Another interesting observation is the ability of {\name} to copy texts from source to predicted text. Even without having an explicit copying mechanism~\cite{see2017get}, {\name} is able to understand the key phrases that co-occur in both Hindi and English, like, numeric and proper nouns and copies these tokens while generating. This shows how our model can also be used in conditional generation of texts. It also ends the sentence with |, which is a common punctuation widely used as a full stop in Hindi texts.

\begin{table*}[ht]
    \centering
    \subfloat[PoS]{
    \resizebox{\textwidth}{!}{
    \begin{tabular}{c|p{25em}|l|p{25em}}
        & \bf Input & Sys & \multicolumn{1}{c}{\bf Prediction} \\ \hline
         \multirow{2}{*}{1} & \textbf{Org:}  \textit{\#surgicalstrike\_X \#pakistan\_X will\_V not\_Neg sleep\_N in\_ADP peace\_N tonight\_N .\_X khamoshi\_V toofan\_V ke\_ADP aane\_V ki\_ADP aahat\_N to\_P nahi\_Neg} & A & \textit{\#surgicalstrike\_X \#pakistan\_X will\_V not\_part\_Neg sleep\_N in\_ADP peace\_N tonight\_N .\_X, khamoshi\_V toofan\_V ke\_ADP aane\_V ki\_ADP aahat\_N to\_P nahi\_part\_neg} \\ 
         & \textbf{Translated:} \textit{\#surgicalstrike \#pakistan will not sleep in peace tonight. Does this silence signify that a storm is approaching} & B & \textit{\#surgicalstrike\_X \#pakistan\_X will\_V not\_part\_Neg sleep\_N in\_ADP peace\_N tonight\_N .\_X, khamoshi\_V toofan\_V ke\_N aane\_V ki\_ADP \textcolor{red}{aahat\_ADP} to\_P nahi\_part\_Neg} \\ \hline 
         \multirow{2}{*}{2} & \textbf{Org:} \textit{minimum\_N cincuenta\_Num mil\_Num por\_ADP persona\_N .\_Punct} & A &  \textit{minimum\_N cincuenta\_Num mil\_Num por\_ADP persona\_N .\_Punct} \\ 
         & \textbf{Translated:} \textit{minimum fifty thousand per person .} & B & \textit{minimum\_N cincuenta\_Num mil\_Num por\_ADP persona\_N .\_Punct} \\ \hline 
    \end{tabular}}
    }
    
      \subfloat[NER]{
      \resizebox{\textwidth}{!}{
    \begin{tabular}{c|p{25em}|l|p{25em}}
         & \bf Input & Sys & \multicolumn{1}{c}{\bf Prediction} \\ \hline
         \multirow{2}{*}{1} & \textbf{Org:} \textit{@gurmeetramrahim \{\underline{dhan dhan satguru}\}$_{Per}$ tera hi aasra \#msgloveshumanity salute 2 \{\underline{msg}\}$_{Org}$ $<$url$>$} & A & \textit{@gurmeetramrahim \{\underline{dhan dhan satguru}\}$_{Per}$ tera hi aasra \#msgloveshumanity salute 2 \textcolor{red}{msg} $<$url$>$} \\ 
         & \textbf{Translated:} & B & \textit{@gurmeetramrahim \{\underline{dhan dhan satguru}\}$_{Per}$ tera hi aasra \#msgloveshumanity salute 2 \{\underline{msg}\}$_{Org}$ $<$url$>$} \\ \hline 
         \multirow{2}{*}{2} & \textbf{Org:} \textit{ste \{\underline{sábado}\}$_{Time}$ nuestras alumnas en \{\underline{imagen modeling}\}$_{Org}$ by \{\underline{la gatita}\}$_{Per}$ reciben la visita de \{\underline{monic abbad}\}$_{Per}$ , joven … $<$url$>$} & A & \textit{ste \{\underline{sábado}\}$_{Time}$ nuestras alumnas en \textcolor{red}{imagen modeling} by \textcolor{red}{la gatita} reciben la visita de \{\underline{monic}\}$_{Per}$ \textcolor{red}{abbad} , joven … $<$url$>$} \\ 
         & \textbf{Translated:} \textit{This saturday our students in image modeling by the kitten receive a visit from young monic abbad} & B & \textit{ste \{\underline{sábado}\}$_{Time}$ nuestras alumnas \textcolor{red}{en imagen} modeling by \textcolor{red}{la gatita} reciben la visita de \{\underline{monic abbad}\}$_{Per}$ , joven … $<$url$>$} \\ \hline 
    \end{tabular}}
    }
    
\subfloat[Sentiment]{
    \resizebox{0.55\textwidth}{!}{
    \begin{tabular}{c|p{25em}|c|c|c}
         & \multirow{2}{*}{\bf Input} & \multirow{2}{*}{\bf Gold} & \multicolumn{2}{c}{\bf Prediction} \\ \cline{4-5}
         & & & \textbf{A} & \bf B \\ \hline
         \multirow{2}{*}{1} & \textbf{Org:} \textit{safal videsh yatra ke liye badhai ho sir} & \multirow{2}{*}{Pos} & \multirow{2}{*}{Pos} & \multirow{2}{*}{\textcolor{red}{Neu}} \\ 
         & \textbf{Trans:} \textit{Congratulations on the successful foreign trip sir} & & & \\ \hline 
         \multirow{2}{*}{2} & \textbf{Org:} \textit{nunca pensé que " bruh " me frustraría tanto} & \multirow{2}{*}{Neu} & \multirow{2}{*}{Neu} & \multirow{2}{*}{\textcolor{red}{Neg}} \\ 
         & \textbf{Trans:} \textit{I never thought that "bruh" would frustrate me so much} & & & \\ \hline 
          \multirow{2}{*}{3} & \textbf{Org:} \textit{desh chodo pahaley yeh media ko change karo ... !! ?} & \multirow{2}{*}{\textcolor{red}{Neu}} & \multirow{2}{*}{\textcolor{red}{Neg}} & \multirow{2}{*}{Neg} \\  
          & \textbf{Trans:} \textit{Leave the country, first change the media} & & & \\ \hline 
    \end{tabular}}
    }   
    \subfloat[MT]{
      \resizebox{0.46\textwidth}{!}{
    \begin{tabular}{c|p{25em}}
    & \\ \hline 
         \multirow{2}{*}{1} & \textbf{Source:} \textit{Licencing and import policies were liberalise} \\
         & \textbf{Reference:} \textit{license tatha import ki policies ko udar banaya gaya} \\ \hdashline 
         & \textbf{\name:} Licencing aur policies liberal the | \\ \hline 
         \multirow{2}{*}{2} & \textbf{Source:} \textit{This fact is based on possibility} \\
         & \textbf{Reference} \textit{yah fact possibility par aadharit hai |} \\ \hdashline
         & \textbf{\name:} yah fact possibility par aadharit hai \\ \hline 
    \end{tabular}}
    }
    \caption{Error Analysis. System A denotes {\name} and B denotes CS-ELMO.}
    \label{tab:error:supp}
\end{table*}

\end{document}